\documentclass[12pt]{article}

\usepackage{amsmath,amsthm, amsfonts, amssymb, amsxtra, amsopn}
\usepackage{pgfplots}
\pgfplotsset{compat=1.13}
\usepackage{pgfplotstable}
\usepackage{graphicx,grffile}
\usepackage{multirow}
\usepackage{algorithm} 
\usepackage[noend]{algpseudocode} 
\usepackage{booktabs}
\usepackage{listings}
\usepackage{cmap}
\usepackage{colortbl}
\usepackage{adjustbox}
\usepackage{epsfig}
\usepackage{mdframed}
\usepackage{tabularx}
\usepackage{longtable}

\usepackage[tableposition=top,font=small,skip=5pt]{caption}
\usepackage{subcaption}
\usepackage{makecell} 

\def\bbb#1{{\color{blue}#1}}

\PassOptionsToPackage{hyphens}{url}

\usepackage{hyperref}
\hypersetup{colorlinks=true,linkcolor=black,citecolor=black,urlcolor=blue,filecolor=black}
\hypersetup{pdfpagemode=UseNone,pdfstartview=}

\usepackage{enumitem}
\setlist[itemize]{noitemsep, topsep=0pt}

\advance\oddsidemargin by -0.35in
\advance\textwidth by 0.7in

\advance\topmargin by -0.4in
\advance\textheight by 0.8in

\long\def\symbolfootnotetext[#1]#2{\begingroup%
\def\thefootnote{\fnsymbol{footnote}}\footnotetext[#1]{#2}\endgroup}

\newcommand\dunderline[3][-1pt]{{%
  \sbox0{#3}%
  \ooalign{\copy0\cr\rule[\dimexpr#1-#2\relax]{\wd0}{#2}}}}
\def\uuu{\kern-1pt\dunderline{0.75pt}{\phantom{M}}}

\title{Temporal Analysis of Adversarial Attacks in Federated Learning}

\author{Rohit Mapakshi\footnotemark[1]\ \ \ 
Sayma Akther\footnotemark[1]\ \ \ 
Mark Stamp\footnotemark[1]\,\,\footnotemark[2]}

\begin{document}

\symbolfootnotetext[1]{Department of Computer Science, San Jose State University}
\symbolfootnotetext[2]{mark.stamp$@$sjsu.edu}

\maketitle

\abstract
In this paper, we experimentally analyze the robustness of selected Federated Learning (FL) 
systems in the presence of adversarial clients.
We find that temporal attacks significantly affect model performance in the FL models tested, 
especially when the adversaries are active throughout or during the later rounds. 
We consider a variety of classic learning models, including Multinominal Logistic Regression (MLR), 
Random Forest, XGBoost, Support Vector Classifier (SVC), as well as various
Neural Network models including Multilayer Perceptron (MLP), 
Convolution Neural Network (CNN), Recurrent Neural Network (RNN), and
Long Short-Term Memory (LSTM). 
Our results highlight the effectiveness of temporal attacks and the need to develop 
strategies to make the FL process more robust against such attacks. 
We also briefly consider the effectiveness of defense mechanisms, 
including outlier detection in the aggregation algorithm.

\section{Introduction}

The rapid evolution in Machine Learning (ML) and the widespread availability of the Internet has made a major 
impact and has become a driving force of technology in numerous fields, including the Internet of Things (IoT), 
Natural Language Processing (NLP), and computer vision. However, machine learning requires a large amount 
of data to train models and typically has operated on centralized data repositories and a centralized server. As more companies 
adopt ML, training models on dispersed data without compromising individual privacy has become 
an important consideration. Data protection laws, such as the 
General Data Protection Regulation (GDPR)~\cite{GDPR2016a}, 
restrict how personal data may be collected and used. 
Federated Learning (FL) has emerged as a powerful solution to data privacy concerns. 

FL leverages the power of distributed and decentralized computing to train ML models. In the FL process, a global model is 
first trained on a subset of the data that is available in a central repository, and the resulting model serves as a starting point. 
This global model 
is then distributed to the clients, who refine the model based on their local data. This allows sensitive user data to be preserved 
because the clients only send the refined model parameters and gradients back to the server. The server then aggregates 
these gradients to optimize the global model and the process repeats. This iterative process continues until the model converges. 
FL not only addresses security and privacy issues, but due to its use of distributed computing, 
it also extends ML capabilities. 

FL systems have tremendous potential in various fields, including healthcare, financial services, recommender systems, and
many others. In healthcare systems, for example, the patient's health records can be kept private while an ML system provides 
personalized health recommendations. In a financial system, the user's spending patterns can be kept private, 
while simultaneously detecting fraudulent activities. IoT devices in cars can be used to monitor traffic and optimize traffic flow 
without compromising an individual driver's privacy. 

There are several inherent challenges in federated learning. Since the client devices vary in terms of computing power, 
storage capacity, and network connectivity, the consistency of model training may be affected. Maintaining consistency 
between low-end smart devices and high-performing servers is a concern in FL. Managing an efficient 
network system to enable data sharing between the FL server and personal device can also be a bottleneck. Furthermore, 
some users may be reluctant to contribute the computing power of their devices for FL systems due to issues such as 
the cost of mobile data transfer and battery consumption. A key issue in FL systems is maintaining the integrity 
of the models developed in the presence of malicious actors within the FL system who might, for example, try to degrade model 
performance by performing poisoning attacks. 

In this paper, we focus on the security aspects of federated learning. Specifically, we aim to measure 
the impact of malicious clients in an FL system, and we quantify the extent to which these adversaries 
can affect the performance of specific FL models. 

This remainder of this paper is organized as follows. Section~\ref{chap:background} discusses the background knowledge 
required to understand federated machine learning. Section~\ref{chap:litreview} discusses relevant previous work in this field. 
Section~\ref{chap:envsetup} explains our experimental setup, and Section~\ref{chap:results} reports our results. 
Finally, Section~\ref{chap:conclusion} provides conclusions drawn from our experiments and we briefly discuss
potential directions for future related work.

\section{Background}\label{chap:background}

In this section, we discuss the fundamentals of Federated Learning (FL), including various types of FL systems 
and some of the 
challenges faced when training models via FL. Then we discuss the different aggregation strategies that are used 
in the FL pipeline. We also briefly consider defense mechanisms, including outlier detection. Finally, we introduce 
the specific classification models that we consider in this paper. 

\subsection{Federated Learning}

Federated learning (FL)~\cite{pmlr-v54-mcmahan17a}, or collaborative learning, is a subfield of machine learning where 
a number of clients work together to train a model while maintaining the decentralization of their data. The fundamental 
idea of FL is to train local models on local data samples of the clients and periodically exchange parameters such as 
weights through a central server. The central server then aggregates these parameters to build a global model. This is not 
the case in typical machine learning environments where data and computing resources are centralized. 

Federated learning differs from distributed learning. In distributed learning, the objective is to parallelize the model 
training process across multiple servers, while the dataset at each client is assumed to be 
Independent and Identically Distributed (IID) 
and roughly the same size. In contrast, in FL, the dataset across clients may be heterogeneous and can range in size by 
orders of magnitude.

\subsubsection{Types of Federated Learning Setups}

\begin{figure}[!htb]
\centering
\includegraphics[scale=0.15]{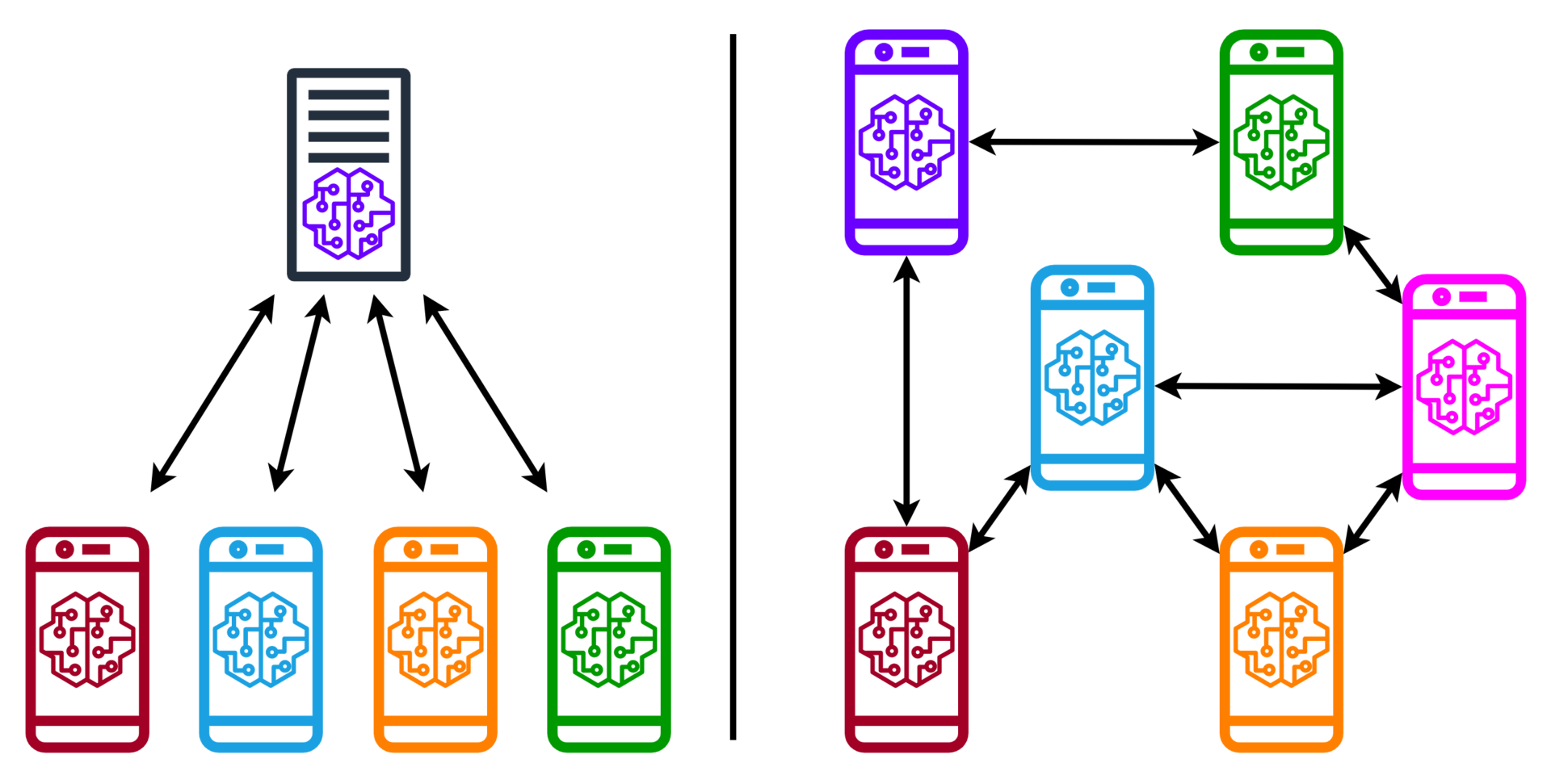}
\caption{Centralized vs decentralized FL~\cite{CentralizedDecentralizedFL2023}}
\label{fig:central}
\end{figure}

Based on the architecture, FL can be centralized, decentralized, or heterogeneous. 
In centralized FL, a central server is responsible for coordinating various steps, such as selecting clients, 
gathering model updates, and aggregating these updates. This setup is prone to a single point of failure at the server. 
In a decentralized FL setup, the clients collaborate among themselves to obtain a global model. This mitigates the problem 
of single-point failures in centralized federated learning. 
In heterogeneous FL, the majority of FL systems assume that local models and global models have the same 
design, but the clients are heterogeneous with varying computing and communication 
capabilities~\cite{diao2020heterofl}.  
For our purposes, the main distinctions between centralized and decentralized FL are illustrated
in Figure~\ref{fig:central}.

According to~\cite{yang2019han}, there are three types of FL systems based on the 
relationship between distributed datasets: Horizontal, Vertical, and Federated Transfer Learning. 
In Horizontal FL, datasets share a similar feature space but differ in sample ID. 
For example, 
suppose two distinct companies are developing a single ML application. They may select similar features, 
but their user data might be different. 
In vertical FL, datasets share the same sample ID space but may have different features. 
For example, the Department of Motor Vehicles and a local city may have overlapping user data but they 
may use different features to develop ML applications that are relevant to their needs.
In Federated Transfer Learning, the datasets are different in both feature space and the sample ID space.


\subsubsection{Centralized Federated Learning Training Process}

A centralized FL training process consists of multiple rounds repeated by a server that coordinates the training process. 
A round typically consists of the following steps.
 
\begin{enumerate}
\item {\bf Client Selection}: The server selects clients that fit certain eligibility criteria. For example, such 
criteria could be based on computing power, connection to unmetered Wi-Fi connection, idleness, etc. 
\item {\bf Broadcast}: The chosen clients download from the server the current ML model and global weights.
\item {\bf Client Computation}: Each client instantiates the training model with the downloaded weights and 
conducts local training on their local dataset. 
\item {\bf Aggregation}: The device updates are aggregated by the server using an aggregating strategy. 
Some client updates may be dropped based on the reliability of the client or aspects of the trained model. 
\item {\bf Model Update}: The aggregated weights are used to reinitialize the global model and the global 
model is evaluated to determine if the FL process has produced an improved model. 
\end{enumerate}

\subsection{Aggregation Strategy}

As mentioned above, the client model weights are aggregated by the server using an aggregating strategy. 
In this section, we discuss two such strategies, namely, federated average and federated bagging.

\subsubsection{Federated Average (FedAvg)}

FedAvg involves computing the average of the shared model weights. 
The intuition is that averaging the gradients amounts to averaging the model weights.
Algorithm~\ref{alg:fedavg} is a FedAvg strategy based on the work in~\cite{pmlr-v54-mcmahan17a}. 

\begin{figure}[!htb]
\algdef{SE}[SUBALG]{Indent}{EndIndent}{}{\algorithmicend\ }%
\algtext*{Indent}
\algtext*{EndIndent}
\centering
\begin{minipage}{0.85\textwidth}
\begin{algorithm}[H]
\caption{FedAvg}\label{alg:fedavg} 
\small
\begin{algorithmic} 
\State \bbb{\texttt{/\!/} $K$ clients indexed by $k$}
\State \bbb{\texttt{/\!/} $\mathcal{P}_k$ is training dataset on client $k$}
\State \bbb{\texttt{/\!/} $n_k=|\mathcal{P}_k|$ and $n=\sum_{k=1}^K n_k$}
\State \bbb{\texttt{/\!/} $B$ is local minibatch size}
\State \bbb{\texttt{/\!/} $E$ is the number of local epochs}
\State \bbb{\texttt{/\!/} $\eta$ is the learning rate}
\State \bbb{\texttt{/\!/} $\ell(w; b)$ is local loss function evaluated on weights~$w$ 
	and minibatch~$b_{\vphantom{\sum_{M_M}}}$}
%
\State \textbf{Server Executes}:
\Indent
   \State initialize $w_0$
   \For{each round $t = 1, 2, \dots$}
     \For{each client $k \in K$ \textbf{in parallel}} \bbb{\texttt{/\!/} all clients update model}
       \State $w_{t+1}^k \gets \Call{ClientUpdate}{k, w_t}$ 
     \EndFor
     \State $w_{t+1} \gets \displaystyle\sum_{k=1}^K \frac{n_k}{n} w_{t+1}^k$ \bbb{\texttt{/\!/} weighted average}
   \EndFor
\EndIndent
\Function{ClientUpdate}{$k, w$} \bbb{\texttt{/\!/} runs on client $k$}
  \State $\mathcal{B} \gets \mbox{(split $\mathcal{P}_k$ into minibatches of size $B$)}$
  \For{each local epoch $i$ from $1$ to $E$}
    \For{each minibatch $b \in \mathcal{B}$}
      \State $w \gets w - \eta \nabla \ell(w; b)$
    \EndFor
  \EndFor
  \State \textbf{return} $w$ to server
\EndFunction
\end{algorithmic}
\end{algorithm}
\end{minipage}%
\end{figure}

\subsubsection{Federated Bagging}

\begin{figure}[!htb]
\centering
\includegraphics[scale=0.25]{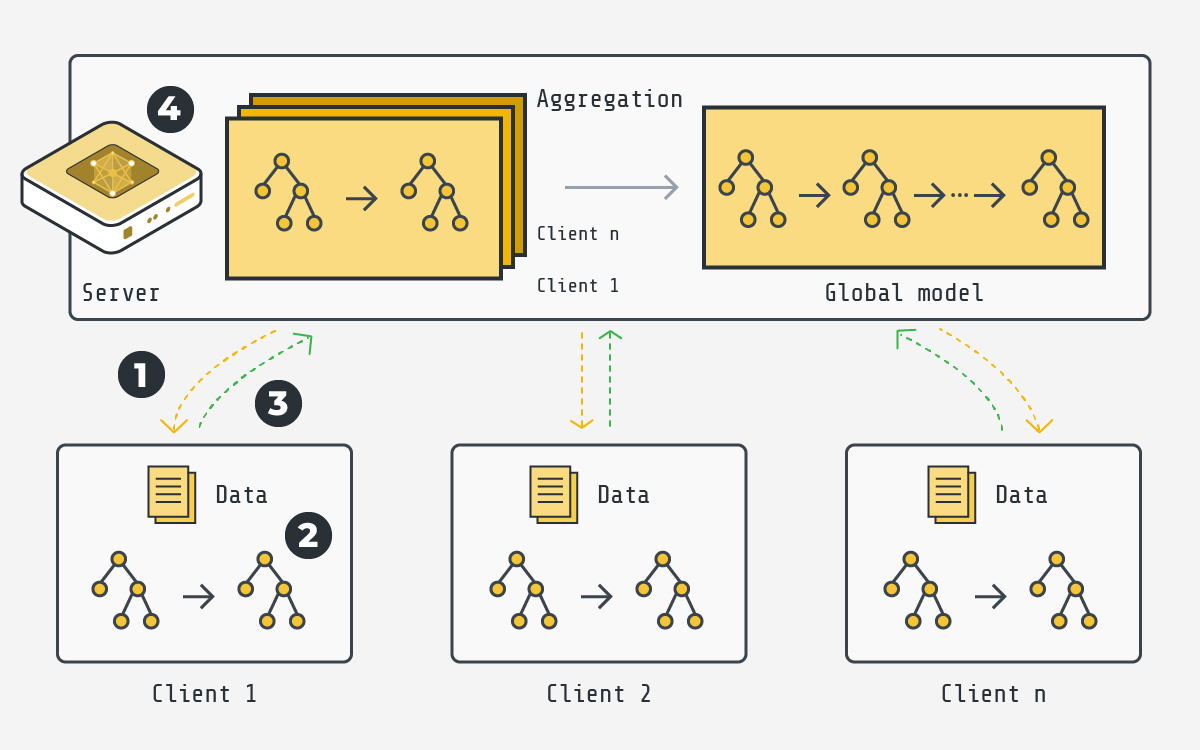}
\caption{Federated bagging~\cite{flowerbagging}}\label{fig:bagging}
\end{figure}

Bagging aggregation~\cite{FlowerAIFedXgbBagging2023} is a technique for generalizing local updates from 
tree-based classifiers, such as Random Forest and XGBoost. Each client is trained on a random subset of the data. 
After every FL round, the server integrates all the trees from the FL clients to form a global model. Therefore, 
all local models affect the global model. For~$M$ clients and~$R$ FL rounds the global model will have a total 
of~$M \times R$ trees. Figure~\ref{fig:bagging} illustrates how federated bagging is performed. 

\subsection{Threats to FL Systems}

Despite its many advantages, FL has significant vulnerabilities due to the distributed nature of such systems. 
The vulnerabilities can be broadly categorized into issues related to client and server integrity, 
as well as general attacks on the distributed nature of FL. Examples of threats to FL models
include the following.

\begin{itemize}
\item \textbf{Compromised Clients}: 
Clients play an active role in model training, which introduces numerous potential vulnerabilities.
For example, malicious clients can send false updates to the server or tamper with the training data, and
thereby corrupt the learning processes.  Also, at the beginning of every FL round, each client receives an 
update from the server. This allows clients to observe intermediate states of the global model, 
and can enable malicious clients to engineer more sophisticated attacks.
\item \textbf{Compromised Server}:
The server in FL coordinates training and model aggregation. A compromised server could 
alter model updates and misdirects the entire learning process. Since the server has access to 
all gradient updates, it might also be able to leak sensitive information about the training data.
\item \textbf{Aggregation Algorithm Vulnerabilities}:
The aggregation algorithm merges client updates. Lack of anomaly detection mechanisms at this stage
could expose the FL system to attacks by the clients. 
\item \textbf{Distributed Nature of FL}:
The decentralized nature of FL allows clients to collude and launch coordinated attacks on the model.
Such attacks may be harder to detect, as compared to attacks by individual clients.
\end{itemize}

\subsection{Specific Attacks on FL}\label{sect:attacks}

For our research, we focus on attacks performed by malicious clients. In general, such
attacks can be broadly categorized as targeted or untargeted.
Targeted attacks aim to alter the model's behavior on specific tasks 
while maintaining overall accuracy on the main task. 
In contrast, untargeted attacks aim to reduce the global model's accuracy;
they do not target specific outcomes.
In this paper, we consider untargeted attacks based on data poisoning, model poisoning,
and GAN reconstruction, as discussed below.

\subsubsection{Data Poisoning Attack}

Data poisoning attacks compromise the integrity of the training data---malicious clients manipulate the 
data in various ways to reduce the overall accuracy of the global model. 
There are two main types of data poisoning attacks, namely, clean-label and dirty label.
In a clean-label attack, the adversarial client manipulates the features of the training data. 
This can be accomplished by adding noise or slightly modifying the training data in such a way that it 
is not easily detected by human evaluation~\cite{cleanlabelattack}.  
In a dirty-label attack, the malicious clients change the labels of the training data. Since the labels are modified, 
the model learning is affected, which can significantly degrade the resulting accuracy~\cite{dirtylabelattack}. 
For our research, we consider dirty-label attacks in an untargeted scenario. 
In our attack, labels are poisoned at a specified percentage
by malicious clients according to
$$
  \mbox{\texttt{poisoned\_label}} = (\mbox{\texttt{original\_label}} + 1)\mbox{ mod $N$}
$$
where~$N$ is the number of classes.

\subsubsection{Model Poisoning Attack}

Model poisoning attacks directly tamper with the model weights and send these malicious weights 
to the server for aggregation. The attack can be altering the gradients sent during the backpropagation phase. 
These attacks can be hard to detect.

Model Poisoning Attacks via Fake Updates (MPAF) was implemented in~\cite{mpaf}. 
This method attacks the learning process using fake updates from malicious clients.
The strategy consists of the following steps. 
\begin{enumerate}
\item A base model with low testing accuracy is used as a starting point. 
\item In each training round where the attack is performed, the client computes the difference 
between the parameters of the base model and the current global model.
\item The malicious clients magnify the difference using a factor $\lambda > 1$.
\end{enumerate}
For our model poisoning attacks, we follow this strategy,
using a randomly initialized model that has the effect of guessing the labels randomly.

\subsubsection{GAN Reconstruction Attack}

Generative Adversarial Network (GAN) is a neural network architecture that can be used to 
generating synthetic data that mimics the training data. GAN includes a
generator network and a discriminator network 
that compete against each other. 
The GAN generator network takes random noise as input and produces fake samples of data. 
Generator training aims to iteratively improve the quality of the fake samples.
The GAN discriminator classifies data as being from the actual dataset or a fake sample 
produced by the generator. The loss is fed back to the generator to improve it. 
GAN training occurs over several rounds in the form of a two-player min-max game. 
In this research, we use Conditional GANs, which enable us to specify a particular
label when training the generator.

The gradients shared for aggregation can reveal features of clients' training data. 
GANs can use this information to create adversarial samples that represent training data. 
In our version of a GAN reconstruction attack, each client has a subset of the classes
that are present in the dataset. A Conditional GAN is used to generate synthetic samples 
of digits that are not present in the local dataset, and we give such samples incorrect labels. 
This has the effect of poisoning the local training in the subsequent rounds, and thereby 
affects the global model.

\subsection{Outlier Detection}

Defense mechanisms can be used in Federated Learning (FL) to mitigate a wide range of attacks 
and to reduce the chance that the global model is corrupted. Outlier detection is a proactive defense mechanism 
that can be enabled in the aggregation stage in FL to identify malicious clients and protect the global 
model from poisoned updates. 


Outlier detection can be viewed as a form of anomaly detection since it serves to detect observations 
that are inconsistent with the rest of the data. This technique is relevant as a defense mechanism in FL,
where it can identify and drop dishonest clients from the FL process. Commonly used outlier detection algorithms 
include the following.

\begin{itemize}
\item \textbf{Robust Covariance}: In this approach, observations are assumed to follow 
a Gaussian distribution, and a robust estimate of the covariance is used to encapsulate the data points 
in an elliptic envelope. All the points that lie outside this envelope are considered to be anomalies~\cite{rousseeuw1999fast}. 
\item \textbf{One-Class SVM}: This techniques consists of training an SVM to learn a decision boundary to separate 
normal points and outliers~\cite{oneclasssvm}. 
\item \textbf{Isolation Forest}: This algorithm repeatedly splits the dataset by randomly selecting features 
and determining a split point between the maximum and minimum values of that feature~\cite{isolationforest}. 
\item \textbf{Local Outlier Factor}: This technique measures the local deviation of data points to 
identify regions of similar density. Since the density is calculated with respect to the neighboring points, 
it can identify anomalies that deviate from the expected pattern~\cite{lof-outlier}. 
\end{itemize}

\subsection{Classification Models}\label{sect:CM}

In machine learning, classification is a task that involves assigning a class label to examples. 
In this paper, we consider classical learning models, neural network models, and ensemble techniques,
all in the context of Federated Learning.

\subsubsection{Multinominal Logistic Regression}

Multinomial Logistic Regression (MLR) is an extension of the logistic regression model
to multiclass problems. For a given set of independent variables, this model predicts the probabilities 
of the possible outcomes for a categorically distributed dependent variable. Logistic regression uses 
maximum-likelihood estimation (MLE) to determine the odd for each class. 

\subsubsection{Support Vector Classifier}

Support Vector Machines (SVM)~\cite{cortes1995support} are popular algorithms for binary classification tasks. 
The algorithm finds a hyperplane that can categorize data points into different classes. The points that 
are closest to the hyperplane are the support vectors and SVMs try to maximize the margin between 
the hyperplane and support vectors. SVMs enable the efficient use of nonlinear decision boundaries via the
so-called kernel trick. Support Vector Classifiers (SVC) extend the SVM
concept to multiclass data.

\subsubsection{Random Forest}

Random Forests~\cite{breiman2001random} is an ensemble learning method that combines multiple 
decision trees and can be used for classification and regression tasks. 
In a decision tree, nodes represent features, branches represent decisions, and the leaf nodes represent an output. 
Decision trees are prone to overfitting and are not effective for data with a large number of features. 
A Random Forest is a collection of decision trees that are independently constructed using 
subsets of the data and feature---a process known as bagging. 
The Random Forest model then uses votes from individual trees for classification. 

\subsubsection{XGBoost}

Extreme Gradient Boosting (XGBoost)~\cite{xgboost}, is a machine learning algorithm used for classification, 
regression, and ranking problems. XGBoost builds upon Gradient-Boosted Decision Trees (GBDT),
which starts with a base decision tree, and makes predictions on the dataset. The errors from this initial prediction 
are used to build the next tree, and his process is repeated iteratively to train the subsequent trees on the residual errors 
of the predecessor. GBDT uses a gradient descent algorithm to minimize the loss between the predicted and 
actual values to minimize the loss function. XGBoost uses a similar approach but constructs trees in parallel 
which significantly improves the computational efficiency of the model. XGBoost also incorporates regularization 
to control overfitting.

\subsubsection{Multilayer Perceptron}

Multilayer Perceptrons (MLP)~\cite{mlp} consists of a series of interconnected nodes or neurons arranged in layers. 
A neuron is an atomic unit that processes incoming signals using a non-linear activation function and then outputs 
a signal. This non-linear activation function enables the network to capture complex data patterns and have made
MLPs a successful model for many classification tasks. 

An MLP includes an input layer, one or more hidden layers, and an output layer.
MLPs are trained in two passes; a forward pass and a backward pass, which together are known as 
backpropagation. In the forward pass, the input data is passed through the network and each layer uses 
the activation function to compute the inputs for the next layer in the network. The backward pass 
is used to propagate the loss backward in the network, effectively adjusting the weights of the neurons
to minimize the loss function. 

\subsubsection{Convolution Neural Network}

Convolutional Neural Networks~\cite{cnn} (CNNs) are a special type of feedforward neural networks that 
are highly effective for image data. These networks are characterized by an input layer, convolution layers, 
pooling layers, and an output layer. 

In each convolutional layer of a CNN, a convolution kernel (or filter), is passed over the input image or 
the outputs of the previous layer. This filter is used to perform a dot product on the data, resulting in a 
map that identifies features, with the features becoming more abstract at each convolutional layer.

Generally, each convolutional layer is followed by a pooling layer, which uses a fixed convolution 
to reduce the size of the generated feature map. There are two common types of pooling: max pooling, 
which takes the maximum value from a group of neurons, and average pooling, which calculates the average value.

The classification step is based on a fully connected layer or multiple such layers. As the name suggests, 
neurons in fully connected layers are connected 
to every neuron in the preceding layer and, if applicable, following layer. 

CNNs do not require any feature engineering as images can be fed directly into the network, including color channels.
Although CNNs were designed for image data, they have proven effective for many types of data that are
not typically considered as images. Any data where local structure dominates is
a good candidate for CNN classifiers.

\subsubsection{Recurrent Neural Networks}

Recurrent Neural Networks (RNNs)~\cite{rnn} are a special type of feedforward neural network that, in
contrast to feedforward networks, can be viewed as having a form of memory. That is, RNNs can capture temporal 
details by retaining information from previous inputs to influence future outputs. RNNs are capable of 
processing sequential data and are highly effective in tasks such as language processing.

When training, RNNs tend to suffer from gradient instability. For example, the gradient can tend zero exponentially
during backpropagation, which severely limits the number of previous time steps that the model can
effectively use, making it difficult to capture long-range dependencies. 

\subsubsection{Long Short-Term Memory}

Long Short-Term Memory (LSTM) models~\cite{lstm} are highly specialized RNNs that are designed
to better deal with long-term dependencies in the data.  While maintaining the structure of RNNs, 
LSTMs include a complex gating structure that improves gradient flow, thereby mitigating
the vanishing and exploding gradient problems that plague generic RNNs.

\section{Literature Review}\label{chap:litreview}

In this literature review of Federated Learning, we first discuss the motivations for 
adopting FL. Next, we consider some of the key challenges in FL, including client dropout, security vulnerabilities, 
and system reliability. Finally, we briefly consider some of the various methods used to evaluate FL systems.

Ensuring data privacy and enabling communication efficiency are the main advantages of FL. 
Data privacy is preserved since training data can remain local~\cite{bonawitz2019federated, reisizadeh2020fedpaq},
while communication efficiency is improved because the local devices (i.e., clients) send only model updates, 
as opposed to the actual data, which would typically incur higher costs for transmission~\cite{pmlr-v54-mcmahan17a}. 
Further, studies show that FL reduces not only network bandwidth but also energy consumption~\cite{9173960}. 
These advantages allow FL systems to scale and attract more clients to participate in the FL process. 

There are some potential disadvantages to FL.
Since the data among the clients may be diverse and heterogeneous, the clients might have data that is
imbalanced and not representative of the feature set for a particular task~\cite{10.1007/978-3-030-20521-8_20}. 
Therefore, training models only on local data can lead to overfitting~\cite{li2020fair}. FL models attempt to overcome 
these issues by collectively aggregating the gradients from multiple clients to create a global model 
that can capture all of the features of a specific dataset. 

Next, we discuss some of the challenges inherent in FL systems. 
These challenges include client dropout, security, reliability,
and system evaluation.

FL process requires multiple rounds of participation by the clients to successfully create a global model,
which increases network bandwidth. In~\cite{8843451} it is claimed that clients tend to drop out of the FL systems 
due to bandwidth limitations, which in turn reduces the amount of data available for model training and 
increases the overall training time. It has been suggested that the server 
avoid aggregating the weights when the number of clients falls below a certain threshold~\cite{liu2020boosting}.

Other research~\cite{8994206} emphasizes selective aggregation based on the quality of the local model,
or asynchronous aggregation~\cite{chen2020asynchronous}. The common goal is
to reduce communication and energy costs while maintaining model performance. 

Incentive mechanisms to client for sharing their resources might 
attract more participants to FL processes~\cite{9006179}. The incentives can be based on 
the quality of the updates provided and the honest behavior of clients. Such incentive mechanisms 
could be orchestrated by a central server~\cite{8851649} or through a distributed blockchain system~\cite{8733825}. 

While the data in an FL system is private to the local devices, there is still a risk of some information being 
exposed via gradient updates. The presence of malicious actors at various levels of an FL system poses a 
significant threat. Various data security mechanisms are considered in~\cite{47246}, 
while \cite{10.1007/978-981-99-7032-2_3} focuses on client device security. 
Not surprisingly, encrypting model updates can help to secure the overall FL system~\cite{Liu2019}. 

Since FL relies on clients participating in the process, it is susceptible to Byzantine attacks,
as discussed in~\cite{choi2019federated}. Auditing mechanisms can be also play a role in 
securing an FL system~\cite{10.1007/978-3-030-35166-3_34}. 

The common problem in any centralized system is that the server is a single point of failure. 
Having a decentralized system mitigates this vulnerability and could make an FL system more reliable. 
Peer-to-Peer approaches for FL have been considered in the literature~\cite{roy2019braintorrent, lalitha2019peertopeer}.  

Apart from handling incentive mechanisms, blockchains can also be used to develop 
data provenance mechanisms to monitor communication between clients to handle 
single-point failures~\cite{8843900}. Further, model updates can be stored in 
Merkle trees~\cite{8892848} to ensure transparent and verifiable records of all transactions in the Fl system. 

In FL, communication efficiency can be measured in terms of communication cost, dropout ratio, 
and system running time~\cite{8945292}. It is also relevant to compare the number of communication 
rounds with learning accuracy~\cite{9139873}, for example.
FL system scalability is evaluated in terms of communication cost and system running time~\cite{8836609}
and overall training time~\cite{10.1145/3338501.3357371}.

\section{Experimental Design}\label{chap:envsetup}

In this section, we first discuss the hardware configuration and the libraries used to implement
our machine learning models. Then we discuss specific detail about our experiments,
including the dataset, FL setup, and the evaluation metrics used. 

\subsection{Hardware and Software}

Table~\ref{tab:hardware} lists the hardware configuration for our experiments. For neural networks models, 
we used PyTorch~\cite{pytorch}, while for machine classic learning models (e.g., Logistic Regression and SVC) 
and outlier detection, we used scikit-learn~\cite{scikit-learn}. For tree-based methods, 
we used the XGBoost~\cite{xgboost-lib} library. For general data processing, we used 
Numpy~\cite{numpy} and pandas~\cite{pandas}.

\begin{table}[!htb]
\centering
\caption{Hardware characteristics}\label{tab:hardware}
\adjustbox{scale=0.85}{
\begin{tabular}{|l|l|}
\hline
\rowcolor[HTML]{C0C0C0} 
\textbf{Feature} & \textbf{Details}                                                        \\ \hline
CPU              & \begin{tabular}[c]{@{}l@{}}AMD Ryzen 5 6600H (3.30 GHz)\end{tabular}    \\ \hline
GPU              & \begin{tabular}[c]{@{}l@{}}NVIDIA GeForce RTX 3060 (6 GB)\end{tabular} \\ \hline
RAM              & 16 GB                                                                   \\ \hline
Storage          & 1 TB                                                                    \\ \hline
\end{tabular}
}
\end{table}

\subsection{Dataset and Data Processing}

\begin{figure}[!htb]
\centering
\includegraphics[scale=0.5]{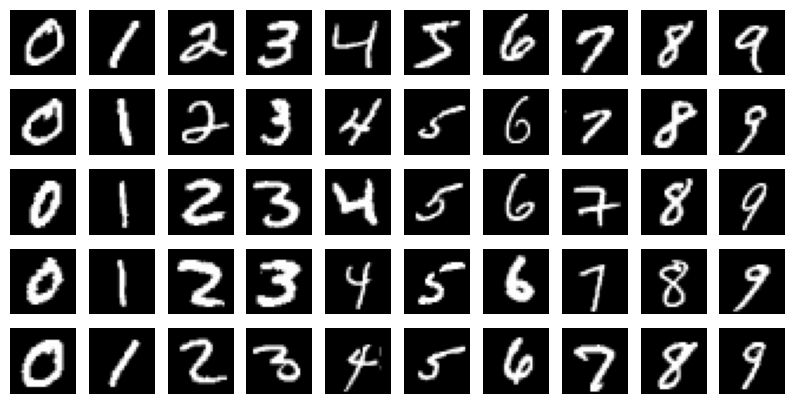}
\caption{Sample MNIST images}
\label{fig:mnist}
\end{figure}

For all of our experiments, we are using the well-known MNIST dataset~\cite{lecun-mnisthandwrittendigit-2010}. 
This dataset consists of a large collection of handwritten digits, 0 through~9, and 
is commonly used as a benchmark for image processing systems. 
MNIST consists of~60,000 training samples and~10,000 test samples. 
All of the samples are in the form of grayscale images of size~$28\times 28$ pixels,
with each pixel value in the range of~0 to~255, where~0 represents black and~255 represents white. 
Examples of images from the dataset are provided in Figure~\ref{fig:mnist}. 

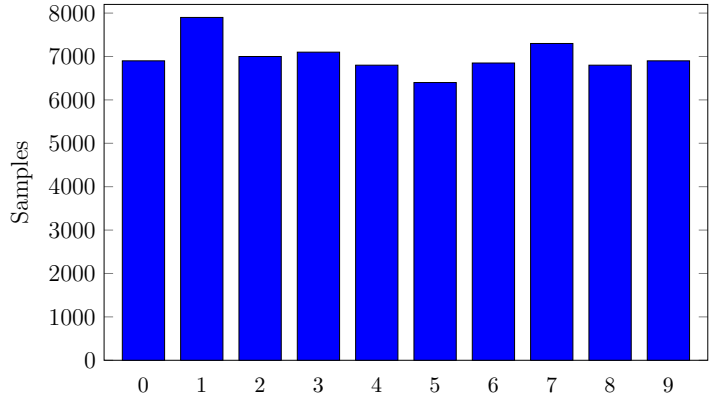
\begin{figure}[!htb]
    \centering
    \begin{tikzpicture}[scale=0.8, every node/.style={scale=0.95}]
\pgfkeys{/pgf/number format/.cd,1000 sep={}}
\begin{axis}[
        width  = 0.75*\textwidth,
        height = 7.5cm,
        ymin=0,ymax=8200,
        ytick={0,1000,2000,3000,4000,5000,6000,7000,8000},
        major x tick style = transparent,
        ybar=5*\pgflinewidth,
        bar width=20.0pt,
        ylabel = {Samples},
        symbolic x coords={0,1,2,3,4,5,6,7,8,9},
        xticklabels={0,1,2,3,4,5,6,7,8,9},
	y tick label style={
    		/pgf/number format/.cd,
   		fixed,
   		fixed zerofill,
    		precision=0},
        xtick = data,
        x tick label style={
		font=\small,
		},
        enlarge x limits=0.075,
        legend cell align=left,
        legend pos=south east,
]
\addplot [fill=blue,opacity=1.00]
coordinates {
(0, 6900)
(1, 7900)
(2, 7000)
(3, 7100)
(4, 6800)
(5, 6400)
(6, 6850)
(7, 7300)
(8, 6800)
(9, 6900)
};
\end{axis}
\end{tikzpicture}
    \caption{Class distribution of MNIST dataset}\label{fig:class_distr}
\end{figure}

The MNIST dataset is approximately balanced across the labels, 0 through~9. The precise
number of samples in each class of the dataset are given in the form of a bar graph in 
Figure~\ref{fig:class_distr}.


As a preprocessing step, the MNIST images are first converted into tensors or numpy arrays, 
depending on the libraries used for the specific classifier. 
The pixel values in the MNIST dataset have a mean of~1.307 and a standard deviation of~0.3081,
and values are normalized to have a mean of~0 and a standard deviation of~1, 
as is standard practice in data preparation. 

For GAN reconstruction attacks, the data is distributed horizontally, in the sense that
the feature space is the same, but only a subset of the classes are present in each client.
Specifically, each partition consists of images for only~7 labels out of~10 classes. 
The conditional GAN generates images of these missing labels and intentionally 
mislabels them with labels that were received by the malicious client.  

\subsection{Federated Learning Setup}\label{sect:FLS}

\begin{figure}[!htb]
\centering
\includegraphics[scale=1.0]{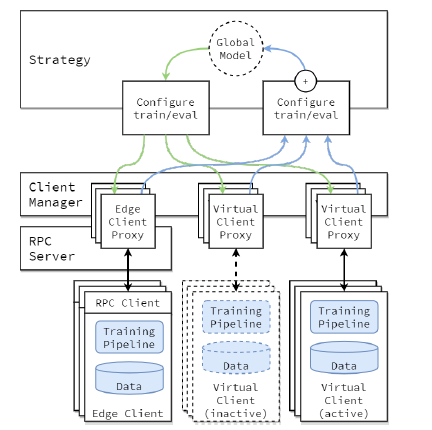}
\caption{Flower federated ML framework \cite{beutel2020flower}}
\label{fig:flower-arch}
\end{figure}

The FL stack developed for this research is based on 
Flower: A Friendly Federated Learning Framework~\cite{beutel2020flower}. 
Flower has three main components, namely, the server, client, and strategy.  
\begin{itemize}
\item {\bf Server}: The Server is responsible for global computations,
including aggregating the model weights, selecting the input parameters for the models, 
and sampling random clients for each FL round. 
\item {\bf Client}: The client is responsible for executing local computations, including running 
the ML model for a set amount of epochs. The client has access to the actual data used for training and evaluation 
of model parameters.
\item {\bf Strategy}: The framework provides a Strategy abstraction which includes the logic for client selection, 
configuration, parameter aggregation, and model evaluation. Outlier detection has been implemented in this 
strategy as a defense mechanism to reject model updates from malicious clients, and is executed on the server. 
A high-level abstraction of the Flower FL framework is provided in Figure~\ref{fig:flower-arch}. 
\end{itemize}

For our experiments, the FedAvg~\cite{pmlr-v54-mcmahan17a} strategy was used to 
aggregate model weights for all models, except that a bagging aggregation strategy~\cite{FlowerAIFedXgbBagging2023} 
was used for aggregating model updates from tree-based models (Random Forest and XGBoost).  
Note that the clients and the server communicate through Remote Procedure Calls (RPC). 

Each experiment was performed for~10 federated rounds and the hyperparameters were adjusted 
accordingly. For example, if a model requires~120 epochs for convergence, the number of local epochs 
is set to~12 in each FL round so that at the end of the FL process, the models would have been trained for 
a total of~120 epochs. 

For LSTM, RNN, Random Forest, and XGBoost, 50 clients were deployed, while for all other
models, 100 clients were deployed.
We set~25\%\ of the clients as adversarial. Recall that we consider three types of attacks, namely, 
untargeted label flipping, model poisoning, and GAN reconstruction.
In this paper, our focus in on temporal effects of adversarial attacks, and hence adversarial clients 
are perform their attacks during different stages of the FL rounds as follows.
\begin{itemize}
    \item \textbf{FULL}: Adversarial attacks are present in all FL rounds
    \item \textbf{MID}: Adversarial attacks are present in~30\% of the middle FL rounds
    \item \textbf{END}: Adversaries attacks are present in the last~30\% of FL rounds
\end{itemize}
Our implementation is specified in detail in Algorithm~\ref{alg:threatmodel}.

\subsection{Evaluation Metrics}

Standard metrics for evaluating classification models include precision, recall, F1-score, loss, and accuracy.
Precision is the ratio of correctly predicted positive observations to the total predicted positive observations.
Recall is the ratio of correctly predicted positive observations to all observations in the actual class.
The F1-Score is computed as the weighted average of Precision and Recall.
Loss is a measure of the error of the model, relative to the specified objective function. 
Accuracy is the ratio of correctly predicted observations to the total observations.
Note that lower values for the loss represent better models and, of course, models with
higher accuracy are desired. We include all of these metrics in our results.

\begin{figure}[!htb]
\algdef{SE}[SUBALG]{Indent}{EndIndent}{}{\algorithmicend\ }%
\algtext*{Indent}
\algtext*{EndIndent}
\centering
\begin{minipage}{0.85\textwidth}
\begin{algorithm}[H]
\caption{Federated learning with threat model}\label{alg:federatedLearningThreat}
\label{alg:threatmodel}
\small
\begin{algorithmic} 
\Procedure{FederatedLearning}{$c, M, e, k, \texttt{attack\_rounds}$}
    \State \textbf{Input}: $c = \mbox{number of clients}$, $M = \mbox{machine learning model}$
    \State \textbf{Input}: $e = \mbox{number of local epochs}$, $k = \mbox{ratio of malicious clients}$
    \State \textbf{Input}: $\texttt{attack\_rounds}\in\{\mbox{FULL, MID, END}\}$
  \State $\texttt{datasets} \gets$ \Call{CreateDistributedDataset}{$c$}
  \State $\texttt{global\_model\_params} \gets \texttt{server}.get\_initial\_params(M)$
  \State $\texttt{clients} \gets$ \Call{SpawnClients}{$\texttt{global\_model\_params}$}
  \State \Call{MarkAdversarialClients}{$\texttt{clients}, k$}
  \For{$i = 1$ to $n$}
    \State \Call{TrainClients}{$\texttt{clients}, \texttt{datasets}, e, i, \texttt{attack\_rounds}$}
    \State $\texttt{server}.\textit{aggregate}(\texttt{clients}.\textit{get\_weights}())$
  \EndFor
  \State $\texttt{global\_model} \gets \texttt{server}.\textit{get\_aggregated\_model}()$
  \State \Call{EvaluateModelOnTestData}{$\texttt{global\_model}$}
\EndProcedure
\Procedure{TrainClients}{$\texttt{clients}, \texttt{datasets}, e, i, \texttt{attack\_rounds}$}
  \For{$j = 1$ to $\textit{length}(\texttt{clients})$}
    \If{$\texttt{clients}[j].\textit{is\_malicious}()$ and $i \in \texttt{attack\_rounds}$}
      \State $\texttt{clients}[j].\textit{perform\_attack}(\texttt{datasets}[j])$
    \EndIf
    \State \Call{Train}{$\texttt{clients}[j], \texttt{datasets}[j], e$}
  \EndFor
\EndProcedure
\end{algorithmic}
\end{algorithm}
\end{minipage}%
\end{figure}

\section{Experiments and Results\label{chap:results}}

In this section, we first present the baseline accuracy for each
of our eight FL models, where baseline refers to the case where
there are no malicious clients.
Then we analyze various outlier detection techniques. Finally, we
turn our attention to experiments for each
of the three types of adversarial attacks discussed in Section~\ref{sect:attacks}, 
namely, a straightforward label flipping attack, a model poisoning attack, and
our GAN reconstruction attack. In each case, 
we consider all of the FL models introduced in Section~\ref{sect:CM} that are relevant
for the particular attack scenario, and we compare the results when no outlier detection
is used to the results obtained when outlier detection is employed.

\subsection{Baseline Cases}

Each of the eight FL models discussed in Section~\ref{sect:CM} was trained via a grid search 
over reasonable sets of hyperparameters. The hyperparameters tested and selected are given
in Appendix~\ref{app:A}. In Figure~\ref{fig:barBase}, we give the accuracy obtained
for each FL model when there are no malicious clients.

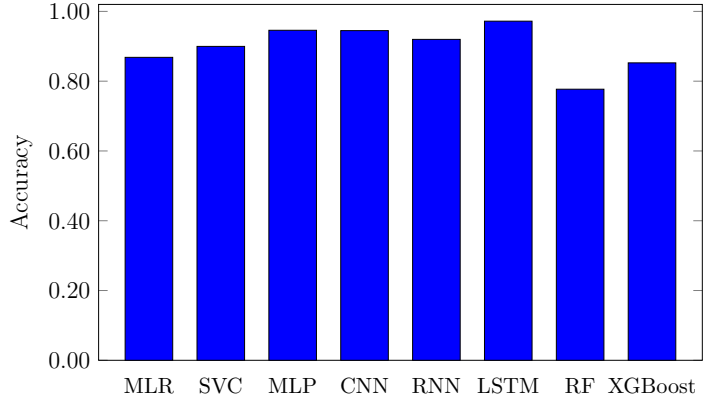
\begin{figure}[!htb]
    \centering
    \begin{tikzpicture}[scale=0.8, every node/.style={scale=0.95}]
\pgfkeys{/pgf/number format/.cd,1000 sep={}}
\begin{axis}[
        width  = 0.75*\textwidth,
        height = 7.5cm,
        ymin=0.0,ymax=1.02,
        ytick={0.0, 0.2, 0.4, 0.6, 0.8, 1.0},
        major x tick style = transparent,
        ybar=5*\pgflinewidth,
        bar width=22.5pt,
        ylabel = {Accuracy},
        symbolic x coords={MLR, SVC, MLP, CNN, RNN,LSTM,RF,XGBoost},
        xticklabels={MLR, SVC, MLP, CNN, RNN,LSTM,RF,XGBoost},
	y tick label style={
    		/pgf/number format/.cd,
   		fixed,
   		fixed zerofill,
    		precision=2},
        xtick = data,
        x tick label style={
		font=\small,
		},
        enlarge x limits=0.10,
        legend cell align=left,
        legend pos=south east,
]
\addplot [fill=blue,opacity=1.00]
coordinates {
(MLR, 0.8683)
(SVC, 0.8997)
(MLP, 0.9459)
(CNN, 0.9449)
(RNN, 0.9198)
(LSTM, 0.9720)
(RF, 0.7770)
(XGBoost, 0.8525)
};
\end{axis}
\end{tikzpicture}
    \caption{Baseline accuracies of FL models}
    \label{fig:barBase}
\end{figure}

From Figure~\ref{fig:barBase} we observe that LSTM achieves the best accuracy, while
MLP and CNN also perform well. In contrast, the tree-based models---Random Forest 
and XGBoost---perform relatively poorly.

\subsection{Outlier Detection Experiments}

We employ a supervised approach to create a classifier that attempts to distinguish between 
honest and malicious clients. First, we train the FL model, as described
in Algorithm~\ref{alg:threatmodel}. In the process, evaluation metrics consisting of
class-wise precision, recall, and F1-scores, and loss are recorded for all clients, whether
honest or malicious. These evaluation metrics 
are each normalized to form a uniform distribution in the interval $[0,1]$. 
The resulting metrics are then used as features to train outlier detectors, 
based on the client labels of honest or malicious. 

\begin{figure}[!htb]
\centering
\includegraphics[scale=0.5]{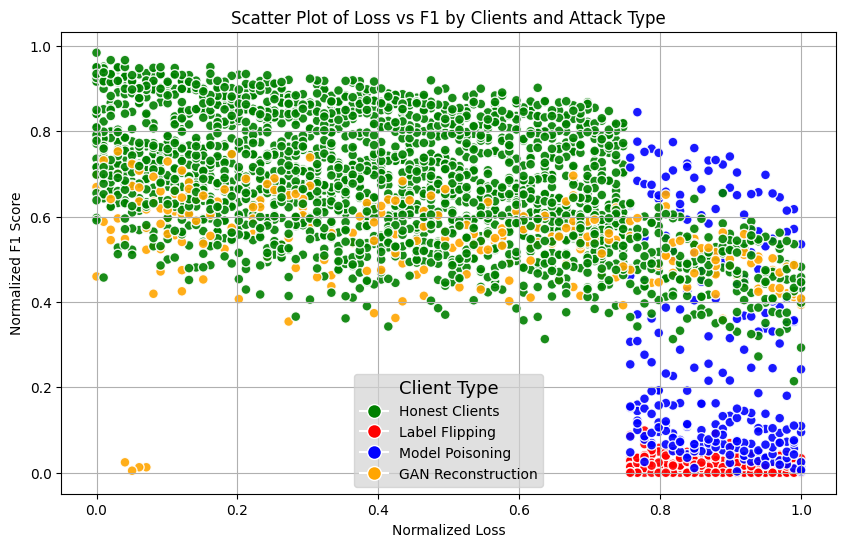}
\caption{Scatterplot of honest and malicious clients}
\label{fig:outlier-distr}
\end{figure}

The distribution of the outlier detection data---in terms of loss and F1-score---is illustrated 
in Figure~\ref{fig:outlier-distr}. Note that the GAN reconstruction data
is similar to the actual data prior to the label flipping that is applied to this data when it is used in an attack.

\begin{figure}[!htb]
\centering
\includegraphics[scale=0.36]{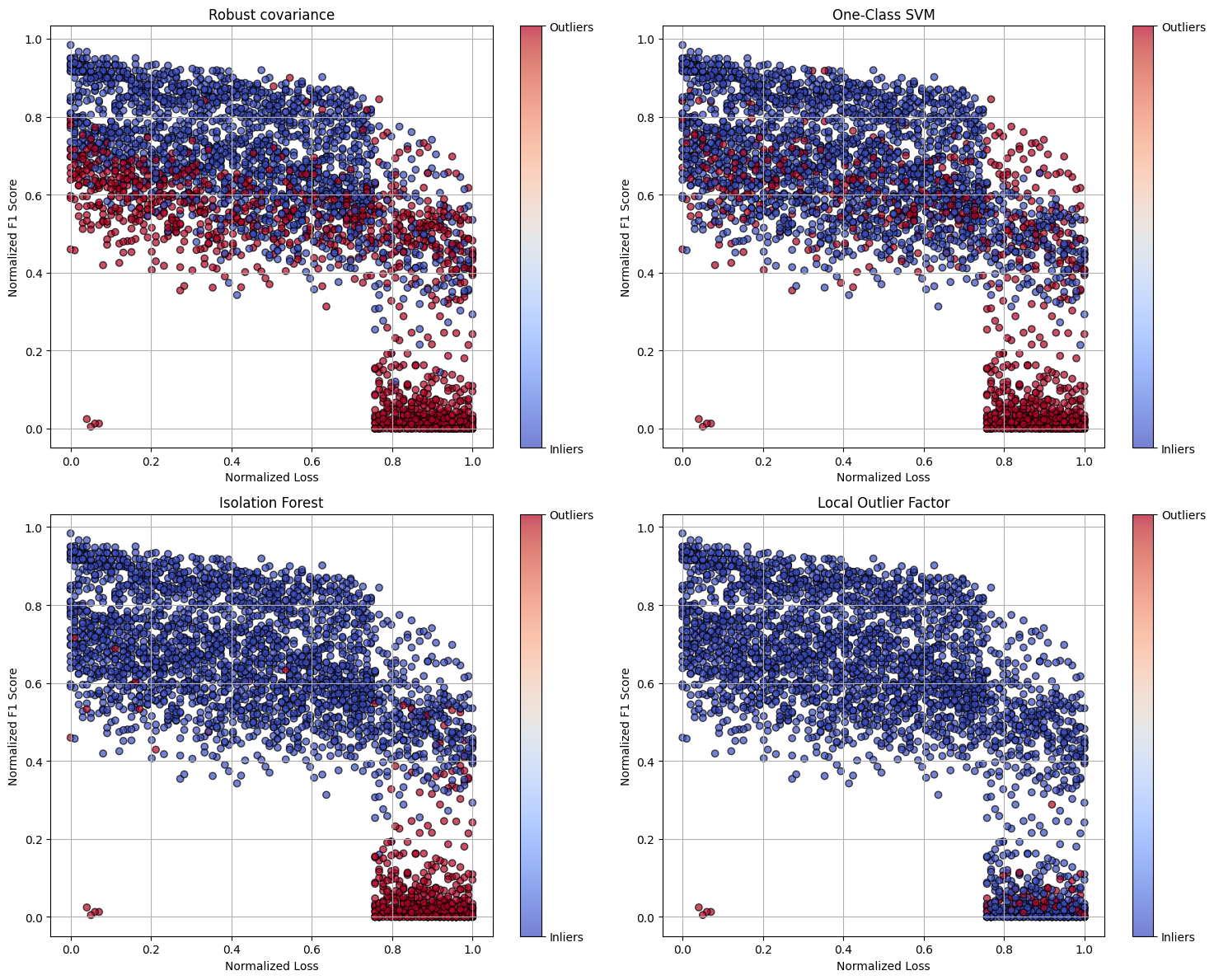}
\caption{Visual comparison of outlier detection algorithms}\label{fig:outlier-comp}
\end{figure}

The following algorithms were tested on our generated outlier dataset: Robust Covariance, 
One-Class SVM, Isolation Forest, and Local Outlier Factor. For each of these methods, 
the best hyperparameters were identified using a grid search. 
The results of this analysis are visualized in Figure~\ref{fig:outlier-comp}.

As a test case, the same set of experiments described in Algorithm~\ref{alg:threatmodel} 
were run again with outlier detection enabled and the results were recorded. 
The accuracy of each of the four tested outlier detection algorithms is given 
in Table~\ref{tab:outlier-acc}. 

\begin{table}[!htb]
\centering
\caption{Accuracy scores for outlier detection algorithms}\label{tab:outlier-acc}
\adjustbox{scale=0.85}{
\begin{tabular}{|l|c|}
\hline
\rowcolor[HTML]{C0C0C0} 
\textbf{Algorithm}            & \textbf{Accuracy} \\ \hline
\textbf{One-Class SVM}        & 0.9700              \\ \hline
\textbf{Isolation Forest}     & 0.8877            \\ \hline
\textbf{Robust Covariance}    & 0.8193            \\ \hline
\textbf{Local Outlier Factor} & 0.8780             \\ \hline
\end{tabular}
}
\end{table}

From the results in Table~\ref{tab:outlier-acc}, we observe that
One-Class SVM far outperforms the other algorithms, 
with an accuracy of~97\%. 
Since the One-Class SVM gives the best results,
in the experiments below, for outlier detection, we employ this technique.

\subsection{Label Flipping Attack}

The accuracies for the temporal test cases FULL, MID, and END
for each of the eight models considered under the label flipping attack
are given in the form of bar graphs in Figures~\ref{fig:barLF}(a), (b), and~(c), respectively.
Table~\ref{tab:lf-metrics-table} in Appendix~\ref{app:B} contain results for all of the metrics 
considered, without outlier detection, while
Table~\ref{tab:lf-metrics-table-outlier} in Appendix~\ref{app:B} gives the analogous results, with
outlier detection enabled. 

\begin{figure}[!htb]
    \centering
    \begin{tabular}{cc}
    \begin{tikzpicture}[scale=0.575, every node/.style={scale=0.95}]
\pgfkeys{/pgf/number format/.cd,1000 sep={}}
\begin{axis}[
        width  = 0.75*\textwidth,
        height = 7.25cm,
        ymin=0.0,ymax=1.02,
        ytick={0.0, 0.2, 0.4, 0.6, 0.8, 1.0},
        major x tick style = transparent,
        ybar=5*\pgflinewidth,
        bar width=12.0pt,
        ylabel = {Accuracy},
        symbolic x coords={MLR, SVC, MLP, CNN, RNN,LSTM,RF,XGBoost},
        xticklabels={MLR, SVC, MLP, CNN, RNN,LSTM,RF,XGBoost},
	y tick label style={
    		/pgf/number format/.cd,
   		fixed,
   		fixed zerofill,
    		precision=2},
        xtick = data,
        x tick label style={
		font=\small,
		},
        enlarge x limits=0.10,
        legend cell align=left,
        legend style={
                at={(0.515,0.2125)},
        },
]
\addplot [fill=blue,opacity=1.00]
coordinates {
(MLR, 0.6187)
(SVC, 0.8707)
(MLP, 0.9240)
(CNN, 0.8805)
(RNN, 0.9155)
(LSTM, 0.9656)
(RF, 0.7240)
(XGBoost, 0.7975)
};
\addlegendentry{Without outlier detection}
\addplot [fill=red,opacity=1.00]
coordinates {
(MLR, 0.7832)
(SVC, 0.8814)
(MLP, 0.9269)
(CNN, 0.9259)
(RNN, 0.9017)
(LSTM, 0.9524)
(RF, 0.7393)
(XGBoost, 0.8360)
};
\addlegendentry{With outlier detection}
\end{axis}
\end{tikzpicture}
    &
    \begin{tikzpicture}[scale=0.575, every node/.style={scale=0.95}]
\pgfkeys{/pgf/number format/.cd,1000 sep={}}
\begin{axis}[
        width  = 0.75*\textwidth,
        height = 7.25cm,
        ymin=0.0,ymax=1.02,
        ytick={0.0, 0.2, 0.4, 0.6, 0.8, 1.0},
        major x tick style = transparent,
        ybar=5*\pgflinewidth,
        bar width=12.0pt,
        ylabel = {Accuracy},
        symbolic x coords={MLR, SVC, MLP, CNN, RNN,LSTM,RF,XGBoost},
        xticklabels={MLR, SVC, MLP, CNN, RNN,LSTM,RF,XGBoost},
	y tick label style={
    		/pgf/number format/.cd,
   		fixed,
   		fixed zerofill,
    		precision=2},
        xtick = data,
        x tick label style={
		font=\small,
		},
        enlarge x limits=0.10,
        legend cell align=left,
        legend style={
                at={(0.515,0.2125)},
        },
]
\addplot [fill=blue,opacity=1.00]
coordinates {
(MLR, 0.8529)
(SVC, 0.8997)
(MLP, 0.9430)
(CNN, 0.9391)
(RNN, 0.9088)
(LSTM, 0.9712)
(RF, 0.7013)
(XGBoost, 0.8436)
};
\addlegendentry{Without outlier detection}
\addplot [fill=red,opacity=1.00]
coordinates {
(MLR, 0.8624)
(SVC, 0.8986)
(MLP, 0.9363)
(CNN, 0.9430)
(RNN, 0.8373)
(LSTM, 0.9708)
(RF, 0.7394)
(XGBoost, 0.8447)
};
\addlegendentry{With outlier detection}
\end{axis}
\end{tikzpicture}
    \\
    \adjustbox{scale=0.85}{(a) FULL}
    &
    \adjustbox{scale=0.85}{(b) MID}
    \\
    \\[-1.5ex]
    \multicolumn{2}{c}{\begin{tikzpicture}[scale=0.575, every node/.style={scale=0.95}]
\pgfkeys{/pgf/number format/.cd,1000 sep={}}
\begin{axis}[
        width  = 0.75*\textwidth,
        height = 7.25cm,
        ymin=0.0,ymax=1.02,
        ytick={0.0, 0.2, 0.4, 0.6, 0.8, 1.0},
        major x tick style = transparent,
        ybar=5*\pgflinewidth,
        bar width=12.0pt,
        ylabel = {Accuracy},
        symbolic x coords={MLR, SVC, MLP, CNN, RNN,LSTM,RF,XGBoost},
        xticklabels={MLR, SVC, MLP, CNN, RNN,LSTM,RF,XGBoost},
	y tick label style={
    		/pgf/number format/.cd,
   		fixed,
   		fixed zerofill,
    		precision=2},
        xtick = data,
        x tick label style={
		font=\small,
		},
        enlarge x limits=0.10,
        legend cell align=left,
        legend style={
                at={(0.515,0.2125)},
        },
]
\addplot [fill=blue,opacity=1.00]
coordinates {
(MLR, 0.7446)
(SVC, 0.8710)
(MLP, 0.9156)
(CNN, 0.7556)
(RNN, 0.9093)
(LSTM, 0.9662)
(RF, 0.6573)
(XGBoost, 0.7402)
};
\addlegendentry{Without outlier detection}
\addplot [fill=red,opacity=1.00]
coordinates {
(MLR, 0.8267)
(SVC, 0.8824)
(MLP, 0.8607)
(CNN, 0.8788)
(RNN, 0.8373)
(LSTM, 0.9534)
(RF, 0.7083)
(XGBoost, 0.7763)
};
\addlegendentry{With outlier detection}
\end{axis}
\end{tikzpicture}}
    \\
    \multicolumn{2}{c}{ \adjustbox{scale=0.85}{(c) END}}
    \end{tabular}
    \caption{Label flipping attack results}\label{fig:barLF}
\end{figure}
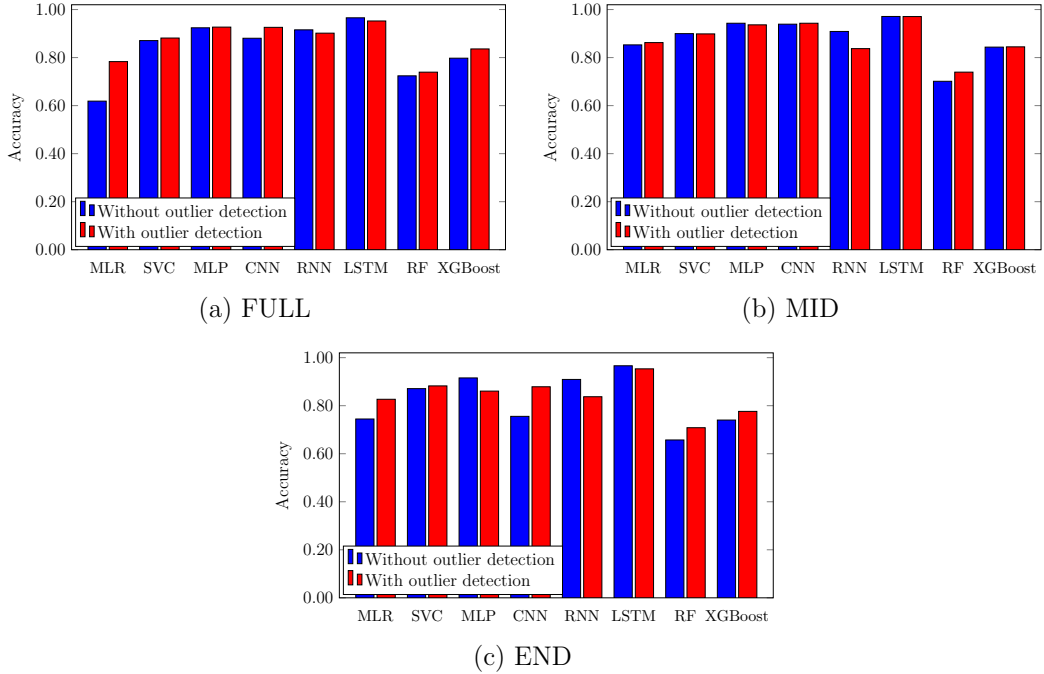

In general, the MID case has little effect on the models, while the END and FULL cases have
a more substantial effect. MLR in the FULL case (and to a lesser extent in the END case), 
and CNN in the END case are affected most by label flipping, and we also observe that 
outlier detection has the largest positive effect in these cases. These results illustrate
the potential benefit of outlier detection in FL.





\subsection{Model Poisoning Attack}

The accuracies for the temporal test cases FULL, MID, and END
for each of the six models\footnote{In model poisoning, the model weight are directly modified. 
Since Random Forest and XGBoost do not have learned weights, 
these models are not included in this section.}
 considered under the model poisoning attack
are given in the form of bar graphs in Figures~\ref{fig:barMP}(a), (b), and~(c), respectively.
Table~\ref{tab:mpaf-metrics-table} in Appendix~\ref{app:B} contain results for all of the metrics 
considered, without outlier detection, while
Table~\ref{tab:mpaf-metrics-table-outlier} in Appendix~\ref{app:B} gives the analogous results, with
outlier detection enabled. 

\begin{figure}[!htb]
    \centering
    \begin{tabular}{cc}
    \begin{tikzpicture}[scale=0.575, every node/.style={scale=0.95}]
\pgfkeys{/pgf/number format/.cd,1000 sep={}}
\begin{axis}[
        width  = 0.75*\textwidth,
        height = 7.25cm,
        ymin=0.0,ymax=1.02,
        ytick={0.0, 0.2, 0.4, 0.6, 0.8, 1.0},
        major x tick style = transparent,
        ybar=5*\pgflinewidth,
        bar width=12.0pt,
        ylabel = {Accuracy},
        symbolic x coords={MLR, SVC, MLP, CNN, RNN,LSTM},
        xticklabels={MLR, SVC, MLP, CNN, RNN,LSTM},
	y tick label style={
    		/pgf/number format/.cd,
   		fixed,
   		fixed zerofill,
    		precision=2},
        xtick = data,
        x tick label style={
		font=\small,
		},
        enlarge x limits=0.10,
        legend cell align=left,
        legend style={
                at={(0.515,0.2125)},
        },
]
\addplot [fill=blue,opacity=1.00]
coordinates {
(MLR, 0.6502)
(SVC, 0.4355)
(MLP, 0.7133)
(CNN, 0.0980)
(RNN, 0.1420)
(LSTM, 0.1427)
};
\addlegendentry{Without outlier detection}
\addplot [fill=red,opacity=1.00]
coordinates {
(MLR, 0.7919)
(SVC, 0.7735)
(MLP, 0.8134)
(CNN, 0.8126)
(RNN, 0.7914)
(LSTM, 0.8357)
};
\addlegendentry{With outlier detection}
\end{axis}
\end{tikzpicture}
    &
    \begin{tikzpicture}[scale=0.575, every node/.style={scale=0.95}]
\pgfkeys{/pgf/number format/.cd,1000 sep={}}
\begin{axis}[
        width  = 0.75*\textwidth,
        height = 7.25cm,
        ymin=0.0,ymax=1.02,
        ytick={0.0, 0.2, 0.4, 0.6, 0.8, 1.0},
        major x tick style = transparent,
        ybar=5*\pgflinewidth,
        bar width=12.0pt,
        ylabel = {Accuracy},
        symbolic x coords={MLR, SVC, MLP, CNN, RNN,LSTM},
        xticklabels={MLR, SVC, MLP, CNN, RNN,LSTM},
	y tick label style={
    		/pgf/number format/.cd,
   		fixed,
   		fixed zerofill,
    		precision=2},
        xtick = data,
        x tick label style={
		font=\small,
		},
        enlarge x limits=0.10,
        legend cell align=left,
        legend style={
                at={(0.515,0.2125)},
        },
]
\addplot [fill=blue,opacity=1.00]
coordinates {
(MLR, 0.6795)
(SVC, 0.8997)
(MLP, 0.8385)
(CNN, 0.1145)
(RNN, 0.1780)
(LSTM, 0.0892)
};
\addlegendentry{Without outlier detection}
\addplot [fill=red,opacity=1.00]
coordinates {
(MLR, 0.7919)
(SVC, 0.8814)
(MLP, 0.9268)
(CNN, 0.9259)
(RNN, 0.9017)
(LSTM, 0.9525)
};
\addlegendentry{With outlier detection}
\end{axis}
\end{tikzpicture}
    \\
    \adjustbox{scale=0.85}{(a) FULL}
    &
    \adjustbox{scale=0.85}{(b) MID}
    \\
    \\[-1.5ex]
    \multicolumn{2}{c}{\begin{tikzpicture}[scale=0.575, every node/.style={scale=0.95}]
\pgfkeys{/pgf/number format/.cd,1000 sep={}}
\begin{axis}[
        width  = 0.75*\textwidth,
        height = 7.25cm,
        ymin=0.0,ymax=1.02,
        ytick={0.0, 0.2, 0.4, 0.6, 0.8, 1.0},
        major x tick style = transparent,
        ybar=5*\pgflinewidth,
        bar width=12.0pt,
        ylabel = {Accuracy},
        symbolic x coords={MLR, SVC, MLP, CNN, RNN,LSTM},
        xticklabels={MLR, SVC, MLP, CNN, RNN,LSTM},
	y tick label style={
    		/pgf/number format/.cd,
   		fixed,
   		fixed zerofill,
    		precision=2},
        xtick = data,
        x tick label style={
		font=\small,
		},
        enlarge x limits=0.10,
        legend cell align=left,
        legend style={
                at={(0.515,0.2125)},
        },
]
\addplot [fill=blue,opacity=1.00]
coordinates {
(MLR, 0.5488)
(SVC, 0.4575)
(MLP, 0.7057)
(CNN, 0.0951)
(RNN, 0.1105)
(LSTM, 0.1525)
};
\addlegendentry{Without outlier detection}
\addplot [fill=red,opacity=1.00]
coordinates {
(MLR, 0.7918)
(SVC, 0.7916)
(MLP, 0.8323)
(CNN, 0.8315)
(RNN, 0.8097)
(LSTM, 0.8552)
};
\addlegendentry{With outlier detection}
\end{axis}
\end{tikzpicture}}
    \\
    \multicolumn{2}{c}{ \adjustbox{scale=0.85}{(c) END}}
    \end{tabular}
    \caption{Model poisoning attack results}\label{fig:barMP}
\end{figure}
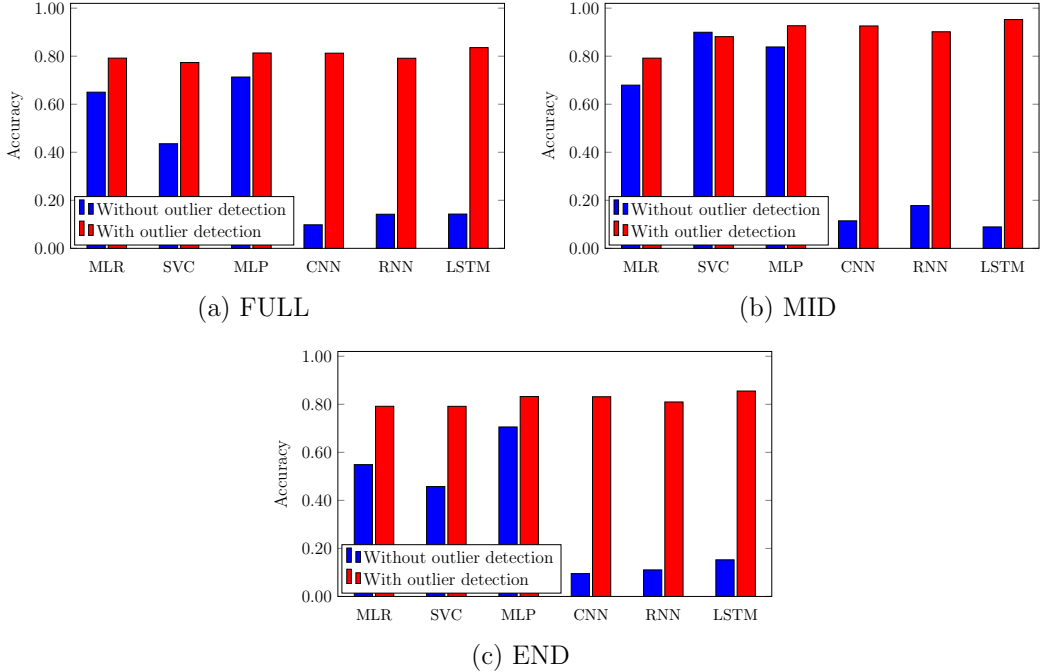


We observe that the model poisoning attack has a drastic effect on the CNN, RNN, and LSTM
models---without outlier detection, these models yield accuracies that are essentially random.
Outlier detection dramatically improves the performance of all of these models, although less
so in the FULL and END cases, as compared to the MID case. With respect to model poisoning,
MLP is the most robust of the models tested.



\subsection{GAN Reconstruction Attack}

The accuracies for the temporal test cases FULL, MID, and END
for each of the eight models considered under the GAN reconstruction attack
are given in the form of bar graphs in Figures~\ref{fig:barGAN}(a), (b), and~(c), respectively.
Table~\ref{tab:gan-metrics-table} in Appendix~\ref{app:B} contain results for all of the metrics 
considered, without outlier detection, while
Table~\ref{tab:gan-metrics-table-outlier} in Appendix~\ref{app:B} gives the analogous results, with
outlier detection enabled. 


Our GAN reconstruction attack is somewhat effective on the tree-based algorithms
of Random Forest and XGBoost, but otherwise the attack has surprisingly little effect. 
Outlier detection has virtually no effect under this attack scenario, with the lone
exception of XGBoost under the END attack scenario.

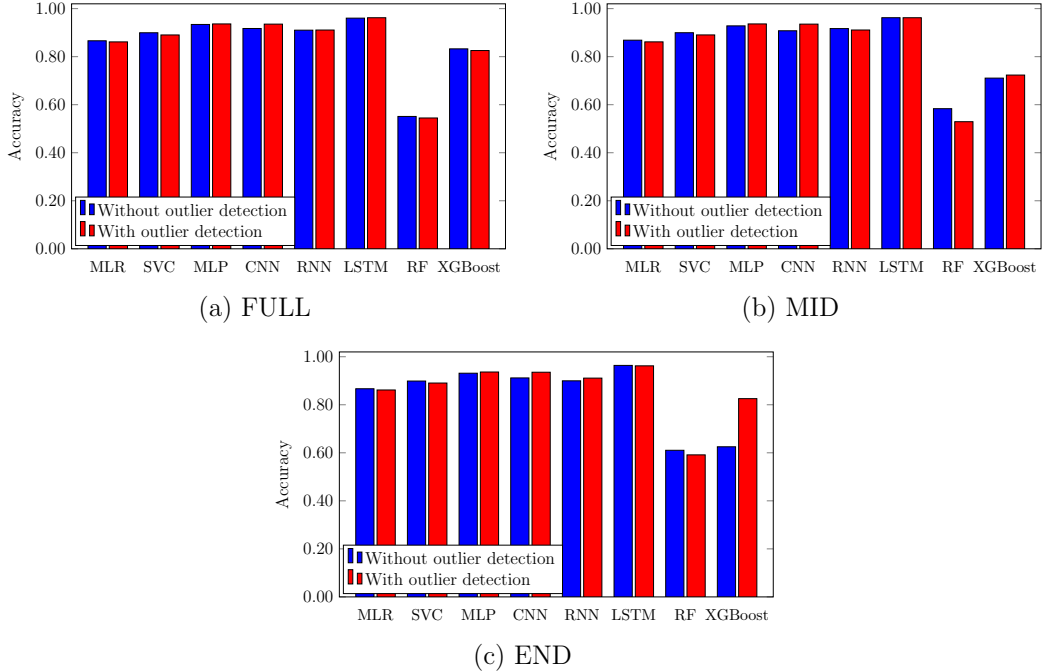
\begin{figure}[!htb]
    \centering
    \begin{tabular}{cc}
    \begin{tikzpicture}[scale=0.575, every node/.style={scale=0.95}]
\pgfkeys{/pgf/number format/.cd,1000 sep={}}
\begin{axis}[
        width  = 0.75*\textwidth,
        height = 7.25cm,
        ymin=0.0,ymax=1.02,
        ytick={0.0, 0.2, 0.4, 0.6, 0.8, 1.0},
        major x tick style = transparent,
        ybar=5*\pgflinewidth,
        bar width=12.0pt,
        ylabel = {Accuracy},
        symbolic x coords={MLR, SVC, MLP, CNN, RNN,LSTM,RF,XGBoost},
        xticklabels={MLR, SVC, MLP, CNN, RNN,LSTM,RF,XGBoost},
	y tick label style={
    		/pgf/number format/.cd,
   		fixed,
   		fixed zerofill,
    		precision=2},
        xtick = data,
        x tick label style={
		font=\small,
		},
        enlarge x limits=0.10,
        legend cell align=left,
        legend style={
                at={(0.515,0.2125)},
        },
]
\addplot [fill=blue,opacity=1.00]
coordinates {
(MLR, 0.8661)
(SVC, 0.8995)
(MLP, 0.9341)
(CNN, 0.9173)
(RNN, 0.9103)
(LSTM, 0.9604)
(RF, 0.5511)
(XGBoost, 0.8325)
};
\addlegendentry{Without outlier detection}
\addplot [fill=red,opacity=1.00]
coordinates {
(MLR, 0.8615)
(SVC, 0.8904)
(MLP, 0.9363)
(CNN, 0.9355)
(RNN, 0.9109)
(LSTM, 0.9621)
(RF, 0.5447)
(XGBoost, 0.8255)
};
\addlegendentry{With outlier detection}
\end{axis}
\end{tikzpicture}
    &
    \begin{tikzpicture}[scale=0.575, every node/.style={scale=0.95}]
\pgfkeys{/pgf/number format/.cd,1000 sep={}}
\begin{axis}[
        width  = 0.75*\textwidth,
        height = 7.25cm,
        ymin=0.0,ymax=1.02,
        ytick={0.0, 0.2, 0.4, 0.6, 0.8, 1.0},
        major x tick style = transparent,
        ybar=5*\pgflinewidth,
        bar width=12.0pt,
        ylabel = {Accuracy},
        symbolic x coords={MLR, SVC, MLP, CNN, RNN,LSTM,RF,XGBoost},
        xticklabels={MLR, SVC, MLP, CNN, RNN,LSTM,RF,XGBoost},
	y tick label style={
    		/pgf/number format/.cd,
   		fixed,
   		fixed zerofill,
    		precision=2},
        xtick = data,
        x tick label style={
		font=\small,
		},
        enlarge x limits=0.10,
        legend cell align=left,
        legend style={
                at={(0.515,0.2125)},
        },
]
\addplot [fill=blue,opacity=1.00]
coordinates {
(MLR, 0.8686)
(SVC, 0.8997)
(MLP, 0.9282)
(CNN, 0.9078)
(RNN, 0.9169)
(LSTM, 0.9622)
(RF, 0.5834)
(XGBoost, 0.7107)
};
\addlegendentry{Without outlier detection}
\addplot [fill=red,opacity=1.00]
coordinates {
(MLR, 0.8616)
(SVC, 0.8905)
(MLP, 0.9363)
(CNN, 0.9355)
(RNN, 0.9110)
(LSTM, 0.9621)
(RF, 0.5292)
(XGBoost, 0.7234)
};
\addlegendentry{With outlier detection}
\end{axis}
\end{tikzpicture}
    \\
    \adjustbox{scale=0.85}{(a) FULL}
    &
    \adjustbox{scale=0.85}{(b) MID}
    \\
    \\[-1.5ex]
    \multicolumn{2}{c}{\begin{tikzpicture}[scale=0.575, every node/.style={scale=0.95}]
\pgfkeys{/pgf/number format/.cd,1000 sep={}}
\begin{axis}[
        width  = 0.75*\textwidth,
        height = 7.25cm,
        ymin=0.0,ymax=1.02,
        ytick={0.0, 0.2, 0.4, 0.6, 0.8, 1.0},
        major x tick style = transparent,
        ybar=5*\pgflinewidth,
        bar width=12.0pt,
        ylabel = {Accuracy},
        symbolic x coords={MLR, SVC, MLP, CNN, RNN,LSTM,RF,XGBoost},
        xticklabels={MLR, SVC, MLP, CNN, RNN,LSTM,RF,XGBoost},
	y tick label style={
    		/pgf/number format/.cd,
   		fixed,
   		fixed zerofill,
    		precision=2},
        xtick = data,
        x tick label style={
		font=\small,
		},
        enlarge x limits=0.10,
        legend cell align=left,
        legend style={
                at={(0.515,0.2125)},
        },
]
\addplot [fill=blue,opacity=1.00]
coordinates {
(MLR, 0.8667)
(SVC, 0.8988)
(MLP, 0.9314)
(CNN, 0.9118)
(RNN, 0.8997)
(LSTM, 0.9639)
(RF, 0.6105)
(XGBoost, 0.6251)
};
\addlegendentry{Without outlier detection}
\addplot [fill=red,opacity=1.00]
coordinates {
(MLR, 0.8616)
(SVC, 0.8905)
(MLP, 0.9364)
(CNN, 0.9355)
(RNN, 0.9110)
(LSTM, 0.9621)
(RF, 0.5914)
(XGBoost, 0.8255)
};
\addlegendentry{With outlier detection}
\end{axis}
\end{tikzpicture}}
    \\
    \multicolumn{2}{c}{ \adjustbox{scale=0.85}{(c) END}}
    \end{tabular}
    \caption{GAN reconstruction attack results}\label{fig:barGAN}
\end{figure}

\section{Conclusion}\label{chap:conclusion}

When adversaries are present in later rounds of the FL process, we tend to observe
a larger negative effect on model performance, while attacks in the earlier rounds do not have
a strong effect. This indicates that FL models can recover from attacks.

For simple attack strategies, outlier detection as a defense mechanism had a clear positive impact,
often significantly improving model performance in the presence of adversarial clients. 
Outlier detection was most effective in the case of model poisoning attacks, which is 
not too surprising, given that this attack strategy was also the most effective. 
The label flipping attack was moderately effective in some cases, while our
GAN reconstruction attacks was surprisingly weak.

Different models showed differing levels of inherent resistance to adversarial attacks. 
For example, although LSTM was the best performing model, it was the most affected by
model poisoning. In contrast, MLP performed almost as well as LSTM in the baseline
case, and yet MLP was the most robust model under the attack scenarios considered.
The ensemble methods of Random Forest and XGBoost struggled 
with GAN Reconstruction attacks, revealing a weakness in tree-based 
when dealing with synthetic adversarial data.

For future work, more sophisticated attack scenarios can be considered. 
Such additional case studies would enable us
to obtain more insight into the relative strengths and weaknesses of the various FL models
analyzed in this paper. Similarly, the effectiveness of more advanced 
defense mechanisms, such as differential privacy---where noise is added to the data to prevent information 
leakage---can be explored. Additionally, instead of a centralized FL scenario, a fully decentralized FL structure 
would be an interesting case study in the context of adversarial attacks.

\bibliographystyle{plain}
\bibliography{references.bib}

\appendix
\makeatletter
\renewcommand\section{\@startsection {section}{1}{\z@}%
                                     {-3.5ex \@plus -1ex \@minus -.2ex}%
                                     {2.3ex \@plus.2ex}%
                                     {\noindent\normalfont\Large\bfseries Appendix }}
\makeatother

\section{}\label{app:A}
\renewcommand{\thesubsection}{A.\arabic{subsection}}
\setcounter{table}{0}
\renewcommand{\thetable}{A.\arabic{table}}
\setcounter{figure}{0}
\renewcommand{\thefigure}{A.\arabic{figure}}


In this appendix, we list the hyperparameters tested (via grid search) for each of the eight 
FL models tested. In each case, we highlight the selected hyperparameters in boldface
and, where appropriate, we specify the model architecture. For each model, we also give the accuracy of 
the trained model in the baseline case, that is, when no adversarial clients are present.

\subsection{MLR and SVC}

For Multinomial Logistic Regression and Support Vector Classifier,
the $28\times28$ images are flattened to a feature vector of~784 features. L2 regularization is 
applied to both models to penalize extreme values. For SVC, a linear kernel is used. 
The hyperparameters considered for logistic regression are in Table~\ref{tab:hp-lgr},
while the hyperparameters for SVC are in Table~\ref{tab:hp-svc}. Note that the 
hyperparameters that yield the best result appear in boldface. In the baseline case,
the MLR model gives an accuracy of~0.8683 while SVC yields an accuracy of~0.8997.

\begin{table}[!htb]
\centering
\caption{Hyperparameters for MLR}\label{tab:hp-lgr}
\adjustbox{scale=0.8}{
\begin{tabular}{|l|l|c|}
\hline
\rowcolor[HTML]{C0C0C0} 
\textbf{Hyperparameter} & \textbf{Values}   & \textbf{Accuracy}        \\ \hline
Local epochs            & {[}\textbf{1}, 10, 100{]} &                          \\ \cline{1-2}
Penalty                 & L2               & \multirow{-2}{*}{0.8683} \\ \hline
\end{tabular}
}
\end{table}

\begin{table}[!htb]
\centering
\caption{Hyperparameters for SVC}\label{tab:hp-svc}
\adjustbox{scale=0.8}{
\begin{tabular}{|l|l|c|}
\hline
\rowcolor[HTML]{C0C0C0} 
\textbf{Hyperparameter} & \textbf{Values}     & \textbf{Accuracy}        \\ \hline
Local epochs            & {[}1, \textbf{10}, 100{]}   &                          \\ \cline{1-2}
C                       & {[}0.01, 0.1, \textbf{1}{]} &                          \\ \cline{1-2}
Kernel                       & \textbf{Linear} &                          \\ \cline{1-2}
Penalty                 & L2                 & \multirow{-4}{*}{0.8997} \\ \hline
\end{tabular}
}
\end{table}


 \subsection{Multilayer Perceptron}

\begin{figure}[!htb]
\centering
\includegraphics[scale=0.285]{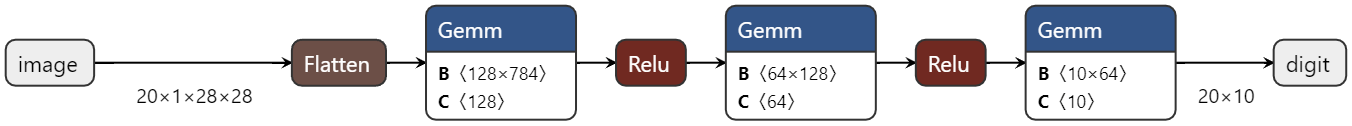}
\caption{MLP architecture}
\label{fig:mlp-arch}
\end{figure}

Figure~\ref{fig:mlp-arch} illustrates our MLP model architecture. The MLP has a flattening layer that 
converts $28\times28$ pixel images into a~784 dimensional vector. This model has three fully connected layers 
with~128, 64, 10 neurons respectively. A ReLU activation function is used after each layer, except the last, 
to introduce nonlinearity. A softmax function is applied to the final layer to convert the probabilities 
to a classification decision. Using on the hyperparameters in boldface in Table~\ref{tab:hp-mlp}, 
this model gives a baseline accuracy of~0.9459
 
\begin{table}[!htb]
\centering
\caption{Hyperparameters for MLP}\label{tab:hp-mlp}
\adjustbox{scale=0.8}{
\begin{tabular}{|l|l|c|}
\hline
\rowcolor[HTML]{C0C0C0} 
\textbf{Hyperparameter} & \textbf{Values}            & \textbf{Accuracy}        \\ \hline
Learning rate           & [\textbf{0.001}, 0.01, 0.1, 1]     & \multirow{4}{*}{0.9459}  \\ \cline{1-2}
Local epochs            & [\textbf{1}, 10, 100]              &                          \\ \cline{1-2}
Batch size              & 20                        &                          \\ \cline{1-2}
Optimizer               & [\textbf{Adam}, SGD]               &                          \\ \hline
\end{tabular}
}
\end{table}

\subsection{Convolution Neural Networks}

\begin{figure}[!htb]
\centering
\includegraphics[scale=0.675]{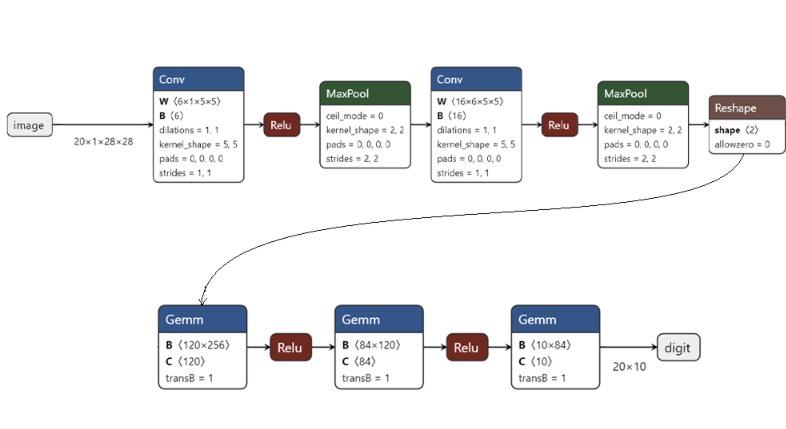}
\caption{CNN architecture}
\label{fig:cnn-arch}
\end{figure}

We use a plain vanilla CNN for classifying digits in the MNIST dataset. The input layer takes an 
image of size $28\times28$. Since the images are gray scale, the number of channels for all 
convolutional layers is~1. The architecture starts with a convolutional layer that applies 
six~$5\times5$ filters to the input. ReLU activation is used to introduce non-linearity. This is 
followed by max pooling to reduce the dimensionality. The second convolutional layer 
applies~16 filters, each of size~$5\times5$, and uses ReLU activation function, 
and is followed by max pooling. Finally, the output from the convolutional layer is flattened and 
passed to a series of fully connected layers of size~120, 84 and~10 neurons, respectively. 
The hyperparameters selected are in boldface in Table~\ref{tab:hp-cnn} and the model architecture 
is illustrated in Figure~\ref{fig:cnn-arch}. For the selected hyperparameters, 
this model achieves an accuracy of~0.9449.

\begin{table}[!htb]
\centering
\caption{Hyperparameters for CNN}\label{tab:hp-cnn}
\adjustbox{scale=0.8}{
\begin{tabular}{|l|l|c|}
\hline
\rowcolor[HTML]{C0C0C0} 
\textbf{Hyperparameter} & \textbf{Values}            & \textbf{Accuracy}        \\ \hline
Learning rate           & {[}0.001, \textbf{0.01}, 0.1, 1{]} &                          \\ \cline{1-2}
Local epochs            & {[}\textbf{1}, 10, 100{]}          &                          \\ \cline{1-2}
Momentum                & 0.9                       &                          \\ \cline{1-2}
Batch size              & 20                        &                          \\ \cline{1-2}
Optimizer               & {[}Adam, \textbf{SGD}{]}           & \multirow{-5}{*}{0.9449} \\ \hline
\end{tabular}
}
\end{table}

\subsection{Recurrent Neural Network}

\begin{figure}[!htb]
\centering
\includegraphics[scale=0.5]{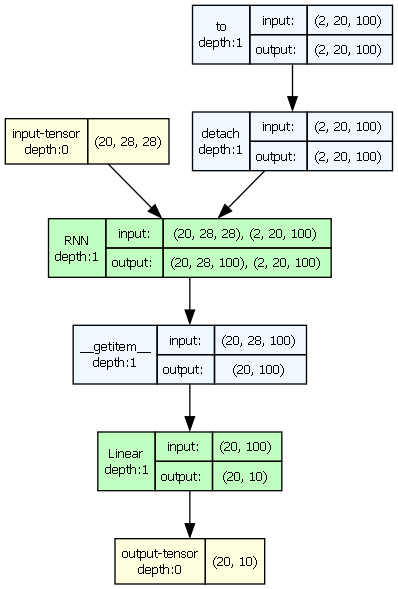}
\caption{RNN architecture}
\label{fig:rnn-arch}
\end{figure}

Our RNN has two cells with~100 neurons. The input to RNN is passed in such a way that 
the~$28\times28$ images are unrolled as~28 sequences of~28 features each. In this way, 
MNIST classification is modeled as sequential data. Tanh activation function is used to introduce non-linearity. 
The last fully connected layer has~10 neurons to represent the~10 digits of MNIST,
with softmax activation function for this last fully connected layer. The hyperparameters 
tested appear in Table~\ref{tab:hp-rnn} and the model architecture is illustrated 
in Figure~\ref{fig:rnn-arch}. For the selected hyperparameters, 
the model has an accuracy of~0.9198. 

\begin{table}[!htb]
\centering
\caption{Hyperparameters for RNN}\label{tab:hp-rnn}
\adjustbox{scale=0.8}{
\begin{tabular}{|l|l|c|}
\hline
\rowcolor[HTML]{C0C0C0} 
\textbf{Hyperparameter} & \textbf{Values}            & \textbf{Accuracy}        \\ \hline
Learning rate           & [0.001, 0.01, \textbf{0.1}, 1]     & \multirow{4}{*}{0.9198}  \\ \cline{1-2}
Local epochs            & [1, \textbf{10}, 100]              &                          \\ \cline{1-2}
Batch size              & 20                        &                          \\ \cline{1-2}
Optimizer               & [Adam, \textbf{SGD}]               &                          \\ \hline
\end{tabular}
}
\end{table}

\subsection{Long Short-Term Memory}

\begin{figure}[!htb]
\centering
\includegraphics[scale=0.5]{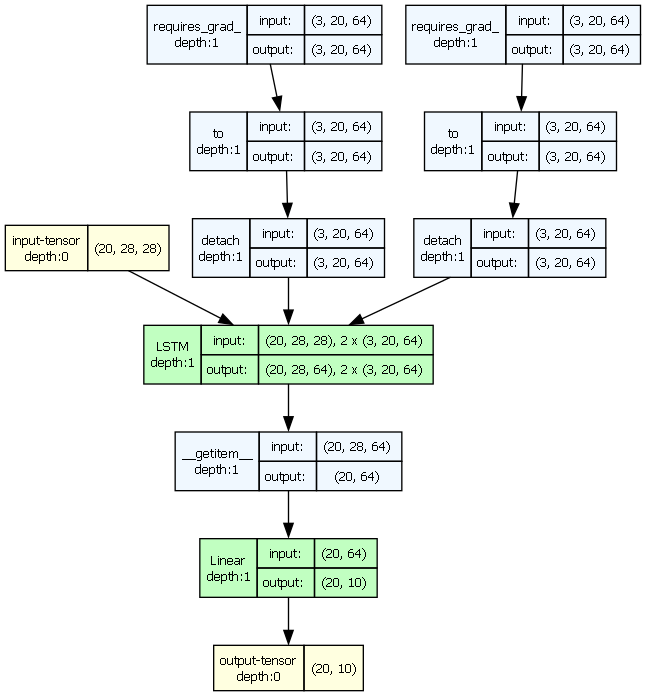}
\caption{LSTM architecture}
\label{fig:lstm-arch}
\end{figure}

Our LSTM architecture is modeled to have three layers, each with~64 hidden states. Here, 
like our RNN, the images are unrolled as~28 sequences of~28 features, thus modeling MNIST 
images as sequential data. The last time step is connected to a fully connected layer of~10 neurons 
representing the~10 MNIST digits. Tanh is used as the activation function in the LSTM layers for 
nonlinearity, and softmax activation is used for the last layer. The hyperparameters tested appear 
in Table~\ref{tab:hp-lstm}, and the LSTM architecture is illustrated in Figure~\ref{fig:lstm-arch}. 
For the selected hyperparameters, the model gave an accuracy of~0.9720. 

\begin{table}[!htb]
\centering
\caption{Hyperparameters for LSTM}\label{tab:hp-lstm}
\adjustbox{scale=0.8}{
\begin{tabular}{|l|l|c|}
\hline
\rowcolor[HTML]{C0C0C0} 
\textbf{Hyperparameter} & \textbf{Values}            & \textbf{Accuracy}        \\ \hline
Learning rate           & [0.001, 0.01, \textbf{0.1}, 1]     & \multirow{4}{*}{0.9720}  \\ \cline{1-2}
Local epochs            & [1, \textbf{10}, 100]              &                          \\ \cline{1-2}
Batch size              & 20                        &                          \\ \cline{1-2}
Optimizer               & [Adam, \textbf{SGD}]               &                          \\ \hline
\end{tabular}
}
\end{table}

\subsection{Random Forest and XGBoost}

For XGBoost, the max depth parameter sets the maximum depth of each decision tree, while
the subsample parameter and colsample by-tree together determine the fraction of features 
to be randomly sampled for each tree. The loss metric is mlogloss due to the multiclass  problem
under consideration. 
The same architecture is used to train a Random Forests by simply setting 
the number of boosting rounds to~1. 
The hyperparameters tested for the Random Forests are in Table~\ref{tab:hp-rf} and
the hyperparameters tested with XGBoost are in Table~\ref{tab:hp-xgb}. 
The Random Forest model achieves an accuracy of~0.770 while XGBoost produces an accuracy of~0.8525.

\begin{table}[!htb]
\centering
\caption{Hyperparameters for Random Forest}\label{tab:hp-rf}
\adjustbox{scale=0.8}{
\begin{tabular}{|l|l|c|}
\hline
\rowcolor[HTML]{C0C0C0} 
\textbf{Hyperparameter} & \textbf{Values}         & \textbf{Accuracy}         \\ \hline
Learning rate     & {[}0.001, \textbf{0.08}, 0.1{]} &                           \\ \cline{1-2}
Max depth               & {[}\textbf{6}, 10, 12{]}        &                           \\ \cline{1-2}
Subsample              & {[}0.50, 0.75, \textbf{0.97}{]} &                           \\ \cline{1-2}
Colsample by-tree       & {[}0.50, 0.75, \textbf{0.97}    &                           \\ \cline{1-2}
Objective               & multi:softmax          &                           \\ \cline{1-2}
Alpha                   & {[}2, 4, \textbf{8}{]}          &                           \\ \cline{1-2}
Lambda                  & {[}\textbf{2}, 4, 8{]}          &                           \\ \cline{1-2}
Evaluation metric       & mlogloss               &                           \\ \cline{1-2}
Tree method             & hist                   &                           \\ \cline{1-2}
Boosting rounds         & 1                      & \multirow{-10}{*}{0.7770} \\ \hline
\end{tabular}
}
\end{table}

\begin{table}[!htb]
\centering
\caption{Hyperparameters for XGBoost}\label{tab:hp-xgb}
\adjustbox{scale=0.8}{
\begin{tabular}{|l|l|c|}
\hline
\rowcolor[HTML]{C0C0C0} 
\textbf{Hyperparameter} & \textbf{Values}         & \textbf{Accuracy}         \\ \hline
Learning rate     & {[}0.001, \textbf{0.08}, 0.1{]} &                           \\ \cline{1-2}
Max depth               & {[}\textbf{6}, 10, 12{]}        &                           \\ \cline{1-2}
Sub sample              & {[}0.50, 0.75, \textbf{0.80}{]} &                           \\ \cline{1-2}
Colsample by-tree       & {[}0.50, 0.75, \textbf{0.80}{]} &                           \\ \cline{1-2}
Objective               & multi:softmax          &                           \\ \cline{1-2}
Alpha                   & {[}2, 4, \textbf{8}{]}          &                           \\ \cline{1-2}
Lambda                  & {[}\textbf{2}, 4, 8{]}          &                           \\ \cline{1-2}
Evaluation metric       & mlogloss               &                           \\ \cline{1-2}
Tree method             & hist                   &                           \\ \cline{1-2}
Boosting rounds         & 60                     & \multirow{-10}{*}{0.8525} \\ \hline
\end{tabular}
}
\end{table}

\clearpage

\section{}\label{app:B}
\renewcommand{\thesubsection}{B.\arabic{subsection}}
\setcounter{table}{0}
\renewcommand{\thetable}{B.\arabic{table}}
\setcounter{figure}{0}
\renewcommand{\thefigure}{B.\arabic{figure}}

In this appendix, we provide tables with all of the metrics for the FULL, MID, and END temporal
cases for each of the three types of attacks considered. For each attack type, we have also included
the baseline case, for comparison.

\subsection{Label Flipping Attack Statistics}

The metrics for our label flipping attacks with no outlier detection enabled are given in Table~\ref{tab:lf-metrics-table}.
The statistics for the corresponding cases with outlier detection enabled are given in Table~\ref{tab:lf-metrics-table-outlier}.

\begin{table}[!htb]
\centering
\caption{Label flipping attack without outlier detection}\label{tab:lf-metrics-table}
\adjustbox{scale=0.6}{
\begin{tabular}{|c|c|c|r|r|r|r|r|r|c|}
\hline
\rowcolor[HTML]{C0C0C0} 
\textbf{}                                                   & \textbf{}          & \textbf{MLR} & \textbf{SVC} & \textbf{MLP} & \textbf{CNN} & \textbf{RNN} & \textbf{LSTM} & \textbf{RF} & \textbf{XGBoost} \\ \hline
\cellcolor[HTML]{C0C0C0}                                    & \textbf{Precision} & 0.8725         & 0.8996       & 0.9459       & 0.9452       & 0.9209       & 0.9720         & 0.7798      & 0.8541           \\ \cline{2-10} 
\cellcolor[HTML]{C0C0C0}                                    & \textbf{Recall}    & 0.8683         & 0.8997       & 0.9459       & 0.9449       & 0.9198       & 0.9720         & 0.7770       & 0.8525           \\ \cline{2-10} 
\cellcolor[HTML]{C0C0C0}                                    & \textbf{F1}        & 0.8669         & 0.8989       & 0.9458       & 0.9448       & 0.9200         & 0.9720         & 0.7760       & 0.8517           \\ \cline{2-10} 
\cellcolor[HTML]{C0C0C0}                                    & \textbf{Loss}      & 0.4984         & 0.9040        & 15.2677      & 13.8814      & 22.3004      & 8.8013        & 1.8893      & 0.6594           \\ \cline{2-10} 
\rowcolor[HTML]{EFEFEF} 
\multirow{-5}{*}{\cellcolor[HTML]{C0C0C0}\textbf{Baseline}} & \textbf{Accuracy}  & 0.8683         & 0.8997       & 0.9459       & 0.9449       & 0.9198       & 0.9720         & 0.7770       & 0.8525           \\ \hline
\cellcolor[HTML]{C0C0C0}                                    & \textbf{Precision} & 0.7233         & 0.8723       & 0.9251       & 0.8858       & 0.9162       & 0.9659        & 0.7252      & 0.7993           \\ \cline{2-10} 
\cellcolor[HTML]{C0C0C0}                                    & \textbf{Recall}    & 0.6187         & 0.8707       & 0.9240        & 0.8805       & 0.9155       & 0.9656        & 0.7240       & 0.7975           \\ \cline{2-10} 
\cellcolor[HTML]{C0C0C0}                                    & \textbf{F1}        & 0.5831         & 0.8700         & 0.9240        & 0.8807       & 0.9154       & 0.9657        & 0.7230       & 0.7978           \\ \cline{2-10} 
\cellcolor[HTML]{C0C0C0}                                    & \textbf{Loss}      & 1.1729         & 1.0891       & 28.7463      & 40.6656      & 24.3846      & 10.2456       & 2.3283      & 1.1623           \\ \cline{2-10} 
\rowcolor[HTML]{EFEFEF} 
\multirow{-5}{*}{\cellcolor[HTML]{C0C0C0}\textbf{FULL}}     & \textbf{Accuracy}  & 0.6187         & 0.8707       & 0.9240        & 0.8805       & 0.9155       & 0.9656        & 0.7240       & 0.7975           \\ \hline
\cellcolor[HTML]{C0C0C0}                                    & \textbf{Precision} & 0.8597         & 0.8996       & 0.9432       & 0.9392       & 0.9095       & 0.9713        & 0.7060       & 0.8460            \\ \cline{2-10} 
\cellcolor[HTML]{C0C0C0}                                    & \textbf{Recall}    & 0.8529         & 0.8997       & 0.9430        & 0.9391       & 0.9088       & 0.9712        & 0.7013      & 0.8436           \\ \cline{2-10} 
\cellcolor[HTML]{C0C0C0}                                    & \textbf{F1}        & 0.8504         & 0.8989       & 0.9430        & 0.9388       & 0.9087       & 0.9712        & 0.6986      & 0.8433           \\ \cline{2-10} 
\cellcolor[HTML]{C0C0C0}                                    & \textbf{Loss}      & 0.6154         & 0.9040        & 15.7765      & 15.4667      & 24.7902      & 9.1022        & 3.2076      & 0.7542           \\ \cline{2-10} 
\rowcolor[HTML]{EFEFEF} 
\multirow{-5}{*}{\cellcolor[HTML]{C0C0C0}\textbf{MID}}      & \textbf{Accuracy}  & 0.8529         & 0.8997       & 0.9430        & 0.9391       & 0.9088       & 0.9712        & 0.7013      & 0.8436           \\ \hline
\cellcolor[HTML]{C0C0C0}                                    & \textbf{Precision} & 0.8124         & 0.8727       & 0.9191       & 0.8166       & 0.9118       & 0.9663        & 0.7003      & 0.7408           \\ \cline{2-10} 
\cellcolor[HTML]{C0C0C0}                                    & \textbf{Recall}    & 0.7446         & 0.8710        & 0.9156       & 0.7556       & 0.9093       & 0.9662        & 0.6573      & 0.7402           \\ \cline{2-10} 
\cellcolor[HTML]{C0C0C0}                                    & \textbf{F1}        & 0.7161         & 0.8703       & 0.9160        & 0.7365       & 0.9092       & 0.9662        & 0.6685      & 0.7398           \\ \cline{2-10} 
\cellcolor[HTML]{C0C0C0}                                    & \textbf{Loss}      & 0.9167         & 1.0849       & 33.2798      & 58.2337      & 25.7339      & 9.9934        & 2.7512      & 1.2845           \\ \cline{2-10} 
\rowcolor[HTML]{EFEFEF} 
\multirow{-5}{*}{\cellcolor[HTML]{C0C0C0}\textbf{END}}      & \textbf{Accuracy}  & 0.7446         & 0.8710        & 0.9156       & 0.7556       & 0.9093       & 0.9662        & 0.6573      & 0.7402           \\ \hline
\end{tabular}%
}
\end{table}

\begin{table}[!htb]
\centering
\caption{Label flipping attack with outlier detection}\label{tab:lf-metrics-table-outlier}
\adjustbox{scale=0.6}{
\begin{tabular}{|c|c|c|r|r|r|r|r|r|c|}
\hline
\rowcolor[HTML]{C0C0C0} 
                                                            &                    & \textbf{MLR} & \textbf{SVC} & \textbf{MLP} & \textbf{CNN} & \textbf{RNN} & \textbf{LSTM} & \textbf{RF} & \textbf{XGBoost} \\ \hline
\cellcolor[HTML]{C0C0C0}                                    & \textbf{Precision} & 0.8725         & 0.8996       & 0.9459       & 0.9452       & 0.9209       & 0.9720         & 0.7798      & 0.8541           \\ \cline{2-10} 
\cellcolor[HTML]{C0C0C0}                                    & \textbf{Recall}    & 0.8683         & 0.8997       & 0.9459       & 0.9449       & 0.9198       & 0.9720         & 0.7770       & 0.8525           \\ \cline{2-10} 
\cellcolor[HTML]{C0C0C0}                                    & \textbf{F1}        & 0.8669         & 0.8989       & 0.9458       & 0.9448       & 0.9200         & 0.9720         & 0.7760       & 0.8517           \\ \cline{2-10} 
\cellcolor[HTML]{C0C0C0}                                    & \textbf{Loss}      & 0.4984         & 0.9040        & 15.2677      & 13.8814      & 22.3004      & 8.8013        & 1.8893      & 0.6594           \\ \cline{2-10} 
\rowcolor[HTML]{EFEFEF} 
\multirow{-5}{*}{\cellcolor[HTML]{C0C0C0}\textbf{Baseline}} & \textbf{Accuracy}  & 0.8683         & 0.8997       & 0.9459       & 0.9449       & 0.9198       & 0.9720         & 0.7770       & 0.8525           \\ \hline
\cellcolor[HTML]{C0C0C0}                                    & \textbf{Precision} & 0.7852         & 0.8816       & 0.9270        & 0.9263       & 0.9024       & 0.9526        & 0.7408      & 0.8370            \\ \cline{2-10} 
\cellcolor[HTML]{C0C0C0}                                    & \textbf{Recall}    & 0.7815         & 0.8817       & 0.9270        & 0.9260        & 0.9014       & 0.9526        & 0.7382      & 0.8354           \\ \cline{2-10} 
\cellcolor[HTML]{C0C0C0}                                    & \textbf{F1}        & 0.7833         & 0.8816       & 0.9270        & 0.9261       & 0.9019       & 0.9526        & 0.7395      & 0.8362           \\ \cline{2-10} 
\cellcolor[HTML]{C0C0C0}                                    & \textbf{Loss}      & 0.6479         & 0.9058       & 22.9016      & 18.0459      & 24.3074      & 8.8189        & 2.8340       & 1.3188           \\ \cline{2-10} 
\rowcolor[HTML]{EFEFEF} 
\multirow{-5}{*}{\cellcolor[HTML]{C0C0C0}\textbf{FULL}}     & \textbf{Accuracy}  & 0.7832         & 0.8814       & 0.9269       & 0.9259       & 0.9017       & 0.9524        & 0.7393      & 0.8360            \\ \hline
\cellcolor[HTML]{C0C0C0}                                    & \textbf{Precision} & 0.8646         & 0.8987       & 0.9365       & 0.9433       & 0.8380        & 0.9711        & 0.7408      & 0.8455           \\ \cline{2-10} 
\cellcolor[HTML]{C0C0C0}                                    & \textbf{Recall}    & 0.8605         & 0.8988       & 0.9364       & 0.9430        & 0.8370        & 0.971         & 0.7382      & 0.8440            \\ \cline{2-10} 
\cellcolor[HTML]{C0C0C0}                                    & \textbf{F1}        & 0.8625         & 0.8987       & 0.9365       & 0.9431       & 0.8375       & 0.9710         & 0.7395      & 0.8448           \\ \cline{2-10} 
\cellcolor[HTML]{C0C0C0}                                    & \textbf{Loss}      & 0.5981         & 0.9492       & 16.0311      & 16.6577      & 24.5304      & 9.2414        & 2.0783      & 0.9891           \\ \cline{2-10} 
\rowcolor[HTML]{EFEFEF} 
\multirow{-5}{*}{\cellcolor[HTML]{C0C0C0}\textbf{MID}}      & \textbf{Accuracy}  & 0.8624         & 0.8986       & 0.9363       & 0.9430        & 0.8373       & 0.9708        & 0.7394      & 0.8447           \\ \hline
\cellcolor[HTML]{C0C0C0}                                    & \textbf{Precision} & 0.8288         & 0.8825       & 0.8608       & 0.8790        & 0.8380        & 0.9536        & 0.7096      & 0.7772           \\ \cline{2-10} 
\cellcolor[HTML]{C0C0C0}                                    & \textbf{Recall}    & 0.8249         & 0.8826       & 0.8608       & 0.8788       & 0.8370        & 0.9535        & 0.7071      & 0.7758           \\ \cline{2-10} 
\cellcolor[HTML]{C0C0C0}                                    & \textbf{F1}        & 0.8269         & 0.8825       & 0.8608       & 0.8789       & 0.8375       & 0.9535        & 0.7084      & 0.7765           \\ \cline{2-10} 
\cellcolor[HTML]{C0C0C0}                                    & \textbf{Loss}      & 0.9968         & 0.9492       & 22.9016      & 62.4664      & 25.6454      & 9.2414        & 2.0783      & 0.9891           \\ \cline{2-10} 
\rowcolor[HTML]{EFEFEF} 
\multirow{-5}{*}{\cellcolor[HTML]{C0C0C0}\textbf{END}}      & \textbf{Accuracy}  & 0.8267         & 0.8824       & 0.8607       & 0.8788       & 0.8373       & 0.9534        & 0.7083      & 0.7763           \\ \hline
\end{tabular}%
}
\end{table}

\clearpage

\subsection{Model Poisoning Attack Statistics}

The metrics for our model poisoning attacks with no outlier detection enabled are given in Table~\ref{tab:mpaf-metrics-table}.
The statistics for the corresponding cases with outlier detection enabled are given in Table~\ref{tab:mpaf-metrics-table-outlier}.

\begin{table}[!htb]
\centering
\caption{Model poisoning attack without outlier detection}\label{tab:mpaf-metrics-table}
\adjustbox{scale=0.6}{
\begin{tabular}{|c|c|c|r|r|r|r|r|}
\hline
\rowcolor[HTML]{C0C0C0} 
\textbf{}                                                   & \textbf{}          & \textbf{MLR} & \textbf{SVC} & \textbf{MLP} & \textbf{CNN} & \textbf{RNN} & \textbf{LSTM} \\ \hline
\cellcolor[HTML]{C0C0C0}                                    & \textbf{Precision} & 0.8725         & 0.8996       & 0.9459       & 0.9452       & 0.9209       & 0.9720         \\ \cline{2-8} 
\cellcolor[HTML]{C0C0C0}                                    & \textbf{Recall}    & 0.8683         & 0.8997       & 0.9459       & 0.9449       & 0.9198       & 0.9720         \\ \cline{2-8} 
\cellcolor[HTML]{C0C0C0}                                    & \textbf{F1}        & 0.8669         & 0.8989       & 0.9458       & 0.9448       & 0.9200         & 0.9720         \\ \cline{2-8} 
\cellcolor[HTML]{C0C0C0}                                    & \textbf{Loss}      & 0.4984         & 0.9040        & 15.2677      & 13.8814      & 22.3004      & 8.8013        \\ \cline{2-8} 
\rowcolor[HTML]{EFEFEF} 
\multirow{-5}{*}{\cellcolor[HTML]{C0C0C0}\textbf{Baseline}} & \textbf{Accuracy}  & 0.8683         & 0.8997       & 0.9459       & 0.9449       & 0.9198       & 0.9720         \\ \hline
\cellcolor[HTML]{C0C0C0}                                    & \textbf{Precision} & 0.6491         & 0.5195       & 0.7135       & 0.0096       & 0.1339       & 0.1673        \\ \cline{2-8} 
\cellcolor[HTML]{C0C0C0}                                    & \textbf{Recall}    & 0.6502         & 0.4355       & 0.7133       & 0.0980        & 0.1420        & 0.1427        \\ \cline{2-8} 
\cellcolor[HTML]{C0C0C0}                                    & \textbf{F1}        & 0.6495         & 0.4219       & 0.7127       & 0.0175       & 0.0680        & 0.1097        \\ \cline{2-8} 
\cellcolor[HTML]{C0C0C0}                                    & \textbf{Loss}      & 11.4170         & 1.8487       & 10784403     & 273.6978     & 478.6425     & 665.9301      \\ \cline{2-8} 
\rowcolor[HTML]{EFEFEF} 
\multirow{-5}{*}{\cellcolor[HTML]{C0C0C0}\textbf{FULL}}     & \textbf{Accuracy}  & 0.6502         & 0.4355       & 0.7133       & 0.0980        & 0.1420        & 0.1427        \\ \hline
\cellcolor[HTML]{C0C0C0}                                    & \textbf{Precision} & 0.7015         & 0.8996       & 0.8381       & 0.1556       & 0.1669       & 0.0925        \\ \cline{2-8} 
\cellcolor[HTML]{C0C0C0}                                    & \textbf{Recall}    & 0.6795         & 0.8997       & 0.8385       & 0.1145       & 0.1780        & 0.0892        \\ \cline{2-8} 
\cellcolor[HTML]{C0C0C0}                                    & \textbf{F1}        & 0.6769         & 0.8989       & 0.8381       & 0.0252       & 0.1547       & 0.0707        \\ \cline{2-8} 
\cellcolor[HTML]{C0C0C0}                                    & \textbf{Loss}      & 2.8480         & 0.9040        & 1202.6790     & 186.6558     & 175.1006     & 424.9485      \\ \cline{2-8} 
\rowcolor[HTML]{EFEFEF} 
\multirow{-5}{*}{\cellcolor[HTML]{C0C0C0}\textbf{MID}}      & \textbf{Accuracy}  & 0.6795         & 0.8997       & 0.8385       & 0.1145       & 0.1780        & 0.0892        \\ \hline
\cellcolor[HTML]{C0C0C0}                                    & \textbf{Precision} & 0.5915         & 0.5608       & 0.7438       & 0.0253       & 0.0780        & 0.1636        \\ \cline{2-8} 
\cellcolor[HTML]{C0C0C0}                                    & \textbf{Recall}    & 0.5488         & 0.4575       & 0.7057       & 0.0951       & 0.1105       & 0.1525        \\ \cline{2-8} 
\cellcolor[HTML]{C0C0C0}                                    & \textbf{F1}        & 0.5305         & 0.4310        & 0.7098       & 0.0170        & 0.0760        & 0.1485        \\ \cline{2-8} 
\cellcolor[HTML]{C0C0C0}                                    & \textbf{Loss}      & 5.1380          & 1.8564       & 3144.8430     & 371.6640      & 614.2485     & 632.3198      \\ \cline{2-8} 
\rowcolor[HTML]{EFEFEF} 
\multirow{-5}{*}{\cellcolor[HTML]{C0C0C0}\textbf{END}}      & \textbf{Accuracy}  & 0.5488         & 0.4575       & 0.7057       & 0.0951       & 0.1105       & 0.1525        \\ \hline
\end{tabular}%
}
\end{table}

\begin{table}[!htb]
\centering
\caption{Model poisoning attack with outlier detection}\label{tab:mpaf-metrics-table-outlier}
\adjustbox{scale=0.6}{
\begin{tabular}{|c|c|c|r|r|r|r|r|}
\hline
\rowcolor[HTML]{C0C0C0} 
                                                            &                    & \textbf{MLR} & \textbf{SVC} & \textbf{MLP} & \textbf{CNN} & \textbf{RNN} & \textbf{LSTM} \\ \hline
\cellcolor[HTML]{C0C0C0}                                    & \textbf{Precision} & 0.8725         & 0.8996       & 0.9459       & 0.9452       & 0.9209       & 0.9720         \\ \cline{2-8} 
\cellcolor[HTML]{C0C0C0}                                    & \textbf{Recall}    & 0.8683         & 0.8997       & 0.9459       & 0.9449       & 0.9198       & 0.9720         \\ \cline{2-8} 
\cellcolor[HTML]{C0C0C0}                                    & \textbf{F1}        & 0.8669         & 0.8989       & 0.9458       & 0.9448       & 0.9200        & 0.9720         \\ \cline{2-8} 
\cellcolor[HTML]{C0C0C0}                                    & \textbf{Loss}      & 0.4984         & 0.9040        & 15.2677      & 13.8814      & 22.3004      & 8.8013        \\ \cline{2-8} 
\rowcolor[HTML]{EFEFEF} 
\multirow{-5}{*}{\cellcolor[HTML]{C0C0C0}\textbf{Baseline}} & \textbf{Accuracy}  & 0.8683         & 0.8997       & 0.9459       & 0.9449       & 0.9198       & 0.9720         \\ \hline
\cellcolor[HTML]{C0C0C0}                                    & \textbf{Precision} & 0.7939         & 0.7736       & 0.8135       & 0.8128       & 0.7919       & 0.8360         \\ \cline{2-8} 
\cellcolor[HTML]{C0C0C0}                                    & \textbf{Recall}    & 0.7902         & 0.7737       & 0.8135       & 0.8126       & 0.7910        & 0.8359        \\ \cline{2-8} 
\cellcolor[HTML]{C0C0C0}                                    & \textbf{F1}        & 0.7920          & 0.7737       & 0.8135       & 0.8127       & 0.7915       & 0.8359        \\ \cline{2-8} 
\cellcolor[HTML]{C0C0C0}                                    & \textbf{Loss}      & 0.5233         & 1.4735       & 15.4204      & 14.7143      & 26.7604      & 9.3294        \\ \cline{2-8} 
\rowcolor[HTML]{EFEFEF} 
\multirow{-5}{*}{\cellcolor[HTML]{C0C0C0}\textbf{FULL}}     & \textbf{Accuracy}  & 0.7919         & 0.7735       & 0.8134       & 0.8126       & 0.7914       & 0.8357        \\ \hline
\cellcolor[HTML]{C0C0C0}                                    & \textbf{Precision} & 0.7939         & 0.8816       & 0.9270        & 0.9263       & 0.9024       & 0.9526        \\ \cline{2-8} 
\cellcolor[HTML]{C0C0C0}                                    & \textbf{Recall}    & 0.7902         & 0.8817       & 0.9270        & 0.9260        & 0.9014       & 0.9526        \\ \cline{2-8} 
\cellcolor[HTML]{C0C0C0}                                    & \textbf{F1}        & 0.7920          & 0.8816       & 0.9270        & 0.9261       & 0.9019       & 0.9526        \\ \cline{2-8} 
\cellcolor[HTML]{C0C0C0}                                    & \textbf{Loss}      & 0.5233         & 1.3831       & 14.6570       & 18.4623      & 30.1055      & 11.7057       \\ \cline{2-8} 
\rowcolor[HTML]{EFEFEF} 
\multirow{-5}{*}{\cellcolor[HTML]{C0C0C0}\textbf{MID}}      & \textbf{Accuracy}  & 0.7919         & 0.8814       & 0.9268       & 0.9259       & 0.9017       & 0.9525        \\ \hline
\cellcolor[HTML]{C0C0C0}                                    & \textbf{Precision} & 0.7939         & 0.7916       & 0.8324       & 0.8317       & 0.8104       & 0.8554        \\ \cline{2-8} 
\cellcolor[HTML]{C0C0C0}                                    & \textbf{Recall}    & 0.7902         & 0.7917       & 0.8324       & 0.8315       & 0.8094       & 0.8554        \\ \cline{2-8} 
\cellcolor[HTML]{C0C0C0}                                    & \textbf{F1}        & 0.7920          & 0.7917       & 0.8324       & 0.8316       & 0.8099       & 0.8554        \\ \cline{2-8} 
\cellcolor[HTML]{C0C0C0}                                    & \textbf{Loss}      & 0.5233         & 1.0938       & 16.4892      & 15.9636      & 25.6454      & 10.1215       \\ \cline{2-8} 
\rowcolor[HTML]{EFEFEF} 
\multirow{-5}{*}{\cellcolor[HTML]{C0C0C0}\textbf{END}}      & \textbf{Accuracy}  & 0.7918         & 0.7916       & 0.8323       & 0.8315       & 0.8097       & 0.8552        \\ \hline
\end{tabular}%
}
\end{table}

\clearpage

\subsection{GAN Reconstruction Attack Statistics}

The metrics for our GAN reconstruction attacks with no outlier detection enabled are given in Table~\ref{tab:gan-metrics-table}.
The statistics for the corresponding cases with outlier detection enabled are given in Table~\ref{tab:gan-metrics-table-outlier}.

\begin{table}[!htb]
\centering
\caption{GAN reconstruction attack without outlier detection}\label{tab:gan-metrics-table}
\adjustbox{scale=0.6}{
\begin{tabular}{|c|c|c|r|r|r|r|r|r|c|}
\hline
\rowcolor[HTML]{C0C0C0} 
                                                            &                    & \textbf{MLR} & \textbf{SVC} & \textbf{MLP} & \textbf{CNN} & \textbf{RNN} & \textbf{LSTM} & \textbf{RF} & \textbf{XGBoost} \\ \hline
\cellcolor[HTML]{C0C0C0}                                    & \textbf{Precision} & 0.8725         & 0.8996       & 0.9459       & 0.9452       & 0.9209       & 0.9720         & 0.7798      & 0.8541           \\ \cline{2-10} 
\cellcolor[HTML]{C0C0C0}                                    & \textbf{Recall}    & 0.8683         & 0.8997       & 0.9459       & 0.9449       & 0.9198       & 0.9720         & 0.7770       & 0.8525           \\ \cline{2-10} 
\cellcolor[HTML]{C0C0C0}                                    & \textbf{F1}        & 0.8669         & 0.8989       & 0.9458       & 0.9448       & 0.9200         & 0.9720         & 0.7760       & 0.8517           \\ \cline{2-10} 
\cellcolor[HTML]{C0C0C0}                                    & \textbf{Loss}      & 0.4984         & 0.9040        & 15.2677      & 13.8814      & 22.3004      & 8.8013        & 1.8893      & 0.6594           \\ \cline{2-10} 
\rowcolor[HTML]{EFEFEF} 
\multirow{-5}{*}{\cellcolor[HTML]{C0C0C0}\textbf{Baseline}} & \textbf{Accuracy}  & 0.8683         & 0.8997       & 0.9459       & 0.9449       & 0.9198       & 0.9720         & 0.7770       & 0.8525           \\ \hline
\cellcolor[HTML]{C0C0C0}                                    & \textbf{Precision} & 0.8721         & 0.8994       & 0.9343       & 0.9183       & 0.9122       & 0.9606        & 0.4128      & 0.8330            \\ \cline{2-10} 
\cellcolor[HTML]{C0C0C0}                                    & \textbf{Recall}    & 0.8661         & 0.8995       & 0.9341       & 0.9173       & 0.9103       & 0.9604        & 0.5511      & 0.8325           \\ \cline{2-10} 
\cellcolor[HTML]{C0C0C0}                                    & \textbf{F1}        & 0.8643         & 0.8987       & 0.9339       & 0.9171       & 0.9104       & 0.9604        & 0.4628      & 0.8318           \\ \cline{2-10} 
\cellcolor[HTML]{C0C0C0}                                    & \textbf{Loss}      & 0.5116         & 0.9114       & 19.4806      & 22.0921      & 24.1009      & 11.4465       & 1.9348      & 0.6909           \\ \cline{2-10} 
\rowcolor[HTML]{EFEFEF} 
\multirow{-5}{*}{\cellcolor[HTML]{C0C0C0}\textbf{FULL}}     & \textbf{Accuracy}  & 0.8661         & 0.8995       & 0.9341       & 0.9173       & 0.9103       & 0.9604        & 0.5511      & 0.8325           \\ \hline
\cellcolor[HTML]{C0C0C0}                                    & \textbf{Precision} & 0.8729         & 0.8996       & 0.9293       & 0.9117       & 0.9181       & 0.9624        & 0.4327      & 0.5942           \\ \cline{2-10} 
\cellcolor[HTML]{C0C0C0}                                    & \textbf{Recall}    & 0.8686         & 0.8997       & 0.9282       & 0.9078       & 0.9169       & 0.9622        & 0.5834      & 0.7107           \\ \cline{2-10} 
\cellcolor[HTML]{C0C0C0}                                    & \textbf{F1}        & 0.8672         & 0.8989       & 0.9281       & 0.9073       & 0.9170        & 0.9622        & 0.4921      & 0.6408           \\ \cline{2-10} 
\cellcolor[HTML]{C0C0C0}                                    & \textbf{Loss}      & 0.5006         & 0.9040        & 19.9489      & 24.1492      & 21.5096      & 10.7564       & 1.9129      & 1.1639           \\ \cline{2-10} 
\rowcolor[HTML]{EFEFEF} 
\multirow{-5}{*}{\cellcolor[HTML]{C0C0C0}\textbf{MID}}      & \textbf{Accuracy}  & 0.8686         & 0.8997       & 0.9282       & 0.9078       & 0.9169       & 0.9622        & 0.5834      & 0.7107           \\ \hline
\cellcolor[HTML]{C0C0C0}                                    & \textbf{Precision} & 0.8718         & 0.8988       & 0.9320        & 0.9127       & 0.9040        & 0.9640         & 0.4546      & 0.4541           \\ \cline{2-10} 
\cellcolor[HTML]{C0C0C0}                                    & \textbf{Recall}    & 0.8667         & 0.8988       & 0.9314       & 0.9118       & 0.8997       & 0.9639        & 0.6105      & 0.6251           \\ \cline{2-10} 
\cellcolor[HTML]{C0C0C0}                                    & \textbf{F1}        & 0.8652         & 0.8980        & 0.9313       & 0.9112       & 0.8996       & 0.9639        & 0.5158      & 0.5216           \\ \cline{2-10} 
\cellcolor[HTML]{C0C0C0}                                    & \textbf{Loss}      & 0.5039         & 0.9119       & 20.064       & 23.1629      & 26.5582      & 10.5914       & 1.9098      & 1.4798           \\ \cline{2-10} 
\rowcolor[HTML]{EFEFEF} 
\multirow{-5}{*}{\cellcolor[HTML]{C0C0C0}\textbf{END}}      & \textbf{Accuracy}  & 0.8667         & 0.8988       & 0.9314       & 0.9118       & 0.8997       & 0.9639        & 0.6105      & 0.6251           \\ \hline
\end{tabular}%
}
\end{table}

\begin{table}[!htb]
\centering
\caption{GAN reconstruction attack with outlier detection}\label{tab:gan-metrics-table-outlier}
\adjustbox{scale=0.6}{
\begin{tabular}{|c|c|c|r|r|r|r|r|r|c|}
\hline
\rowcolor[HTML]{C0C0C0} 
                                                            &                    & \textbf{MLR} & \textbf{SVC} & \textbf{MLP} & \textbf{CNN} & \textbf{RNN} & \textbf{LSTM} & \textbf{RF} & \textbf{XGBoost} \\ \hline
\cellcolor[HTML]{C0C0C0}                                    & \textbf{Precision} & 0.8725         & 0.8996       & 0.9459       & 0.9452       & 0.9209       & 0.9720         & 0.7798      & 0.8541           \\ \cline{2-10} 
\cellcolor[HTML]{C0C0C0}                                    & \textbf{Recall}    & 0.8683         & 0.8997       & 0.9459       & 0.9449       & 0.9198       & 0.9720         & 0.7770       & 0.8525           \\ \cline{2-10} 
\cellcolor[HTML]{C0C0C0}                                    & \textbf{F1}        & 0.8669         & 0.8989       & 0.9458       & 0.9448       & 0.9200         & 0.9720         & 0.7760       & 0.8517           \\ \cline{2-10} 
\cellcolor[HTML]{C0C0C0}                                    & \textbf{Loss}      & 0.4984         & 0.9040        & 15.2677      & 13.8814      & 22.3004      & 8.8013        & 1.8893      & 0.6594           \\ \cline{2-10} 
\rowcolor[HTML]{EFEFEF} 
\multirow{-5}{*}{\cellcolor[HTML]{C0C0C0}\textbf{Baseline}} & \textbf{Accuracy}  & 0.8683         & 0.8997       & 0.9459       & 0.9449       & 0.9198       & 0.9720         & 0.7770       & 0.8525           \\ \hline
\cellcolor[HTML]{C0C0C0}                                    & \textbf{Precision} & 0.8637         & 0.8906       & 0.9365       & 0.9357       & 0.9117       & 0.9623        & 0.5459      & 0.8264           \\ \cline{2-10} 
\cellcolor[HTML]{C0C0C0}                                    & \textbf{Recall}    & 0.8596         & 0.8907       & 0.9364       & 0.9355       & 0.9106       & 0.9623        & 0.5439      & 0.8249           \\ \cline{2-10} 
\cellcolor[HTML]{C0C0C0}                                    & \textbf{F1}        & 0.8617         & 0.8906       & 0.9365       & 0.9356       & 0.9111       & 0.9623        & 0.5449      & 0.8256           \\ \cline{2-10} 
\cellcolor[HTML]{C0C0C0}                                    & \textbf{Loss}      & 0.5283         & 0.9402       & 15.8784      & 14.4367      & 20.5163      & 8.0972        & 2.135       & 0.7583           \\ \cline{2-10} 
\rowcolor[HTML]{EFEFEF} 
\multirow{-5}{*}{\cellcolor[HTML]{C0C0C0}\textbf{FULL}}     & \textbf{Accuracy}  & 0.8615         & 0.8904       & 0.9363       & 0.9355       & 0.9109       & 0.9621        & 0.5447      & 0.8255           \\ \hline
\cellcolor[HTML]{C0C0C0}                                    & \textbf{Precision} & 0.8637         & 0.8906       & 0.9365       & 0.9357       & 0.9117       & 0.9623        & 0.5303      & 0.7243           \\ \cline{2-10} 
\cellcolor[HTML]{C0C0C0}                                    & \textbf{Recall}    & 0.8596         & 0.8907       & 0.9364       & 0.9355       & 0.9106       & 0.9623        & 0.5284      & 0.7229           \\ \cline{2-10} 
\cellcolor[HTML]{C0C0C0}                                    & \textbf{F1}        & 0.8617         & 0.8906       & 0.9365       & 0.9356       & 0.9111       & 0.9623        & 0.5293      & 0.7236           \\ \cline{2-10} 
\cellcolor[HTML]{C0C0C0}                                    & \textbf{Loss}      & 0.5184         & 0.9582       & 14.6570       & 13.3262      & 23.6384      & 9.3294        & 1.9838      & 1.4125           \\ \cline{2-10} 
\rowcolor[HTML]{EFEFEF} 
\multirow{-5}{*}{\cellcolor[HTML]{C0C0C0}\textbf{MID}}      & \textbf{Accuracy}  & 0.8616         & 0.8905       & 0.9363       & 0.9355       & 0.9110        & 0.9621        & 0.5292      & 0.7234           \\ \hline
\cellcolor[HTML]{C0C0C0}                                    & \textbf{Precision} & 0.8637         & 0.8906       & 0.9365       & 0.9357       & 0.9117       & 0.9623        & 0.5927      & 0.8264           \\ \cline{2-10} 
\cellcolor[HTML]{C0C0C0}                                    & \textbf{Recall}    & 0.8596         & 0.8907       & 0.9364       & 0.9355       & 0.9106       & 0.9623        & 0.5905      & 0.8249           \\ \cline{2-10} 
\cellcolor[HTML]{C0C0C0}                                    & \textbf{F1}        & 0.8617         & 0.8906       & 0.9365       & 0.9356       & 0.9111       & 0.9623        & 0.5916      & 0.8256           \\ \cline{2-10} 
\cellcolor[HTML]{C0C0C0}                                    & \textbf{Loss}      & 0.4785         & 0.9402       & 14.0463      & 14.7143      & 23.1924      & 9.1534        & 1.9082      & 1.4125           \\ \cline{2-10} 
\rowcolor[HTML]{EFEFEF} 
\multirow{-5}{*}{\cellcolor[HTML]{C0C0C0}\textbf{END}}      & \textbf{Accuracy}  & 0.8616         & 0.8905       & 0.9364       & 0.9355       & 0.9110        & 0.9621        & 0.5914      & 0.8255           \\ \hline
\end{tabular}%
}
\end{table}

%
%
%
%

\end{document}